\documentclass[preprint,12pt]{elsarticle}

\usepackage{amssymb}
\usepackage{amsmath}
\usepackage{algorithmic}
\usepackage{algorithm}
\usepackage{multirow}
\usepackage[caption=false,font=normalsize,labelfont=sf,textfont=sf]{subfig}
\usepackage{tabularx}
\usepackage{url}

\newtheorem{definition}{Definition}

\journal{Pattern Recognition}

\begin{document}

\begin{frontmatter}



\title{An Efficient Loop and Clique Coarsening Algorithm for Graph Classification}


\author{Xiaorui Qi, Qijie Bai}
\author{Yanlong Wen\corref{cor1}}
\author[l1]{Haiwei Zhang, Xiaojie Yuan} 
\cortext[cor1]{Corresponding author}
\affiliation[l1]{organization={College of Computer Science, Nankai University}, 
    country={China}}

\begin{abstract}
Graph Transformers (GTs) have made remarkable achievements in graph-level tasks.
However, most existing works regard graph structures as a form of guidance or bias for enhancing node representations, which focuses on node-central perspectives and lacks explicit representations of edges and structures.
One natural question arises as to whether we can leverage a hypernode to represent some structures.
Through experimental analysis, we explore the feasibility of this assumption.
Based on our findings, we propose an efficient \textbf{L}oop and \textbf{C}lique \textbf{C}oarsening algorithm with linear complexity \textbf{for} \textbf{G}raph \textbf{C}lassification (LCC4GC) on GT architecture.
Specifically, we build three unique views, original, coarsening, and conversion, to learn a thorough structural representation.
We compress loops and cliques via hierarchical heuristic graph coarsening and restrict them with well-designed constraints, which builds the coarsening view to learn high-level interactions between structures.
We also introduce line graphs for edge embeddings and switch to edge-central perspective to alleviate the impact of coarsening reduction.
Experiments on eight real-world datasets demonstrate the improvements of LCC4GC over 31 baselines from various architectures.
Our code is available: \url{https://github.com/NickSkyyy/LCC-for-GC}.
\end{abstract}




\end{frontmatter}


\section{Introduction}
Graph Transformers (GTs) have succeeded in graph classification tasks~\cite{ref:GraphGPS,ref:SA-GAT,ref:UGT}.
GTs utilize position encoding to represent the graph topology globally, jumping out of the neighborhood restriction and interacting between distant node pairs.
Although GTs can learn long-distance structural information, most GNNs and GTs use graph structures as guidance~\cite{ref:T2GNN} or bias~\cite{ref:Graphormer} to obtain better node representations rather than directly representing them.

Structural information is crucial for graph-level tasks, urging Graph Structural Learning (GSL)~\cite{ref:NIPS21,ref:RDGSL,ref:TKDE24}.
GSL aims to learn the optimized graph structure and representation from the original graph, further improving the model performance on downstream tasks (e.g. graph classification).
To fully utilize topological information, one natural question arises as to whether we can leverage a hypernode to represent some structures.
Take the molecule graph as an example, specific structures such as molecular fragments and functional groups contain rich semantics.
When we coarsen these structures into hypernodes, the new graph contains the interaction information with molecule fragments and functional groups as basic units, called high-level structure.
Research~\cite{ref:DIFFPOOL} has proved that models can benefit from them.
Meanwhile, random perturbation of these structures will produce additional structural information, which has been proved to be invalid nevertheless~\cite{ref:MolHF}.
\begin{figure*}[!t]
    \centering
    \includegraphics[width=\linewidth]{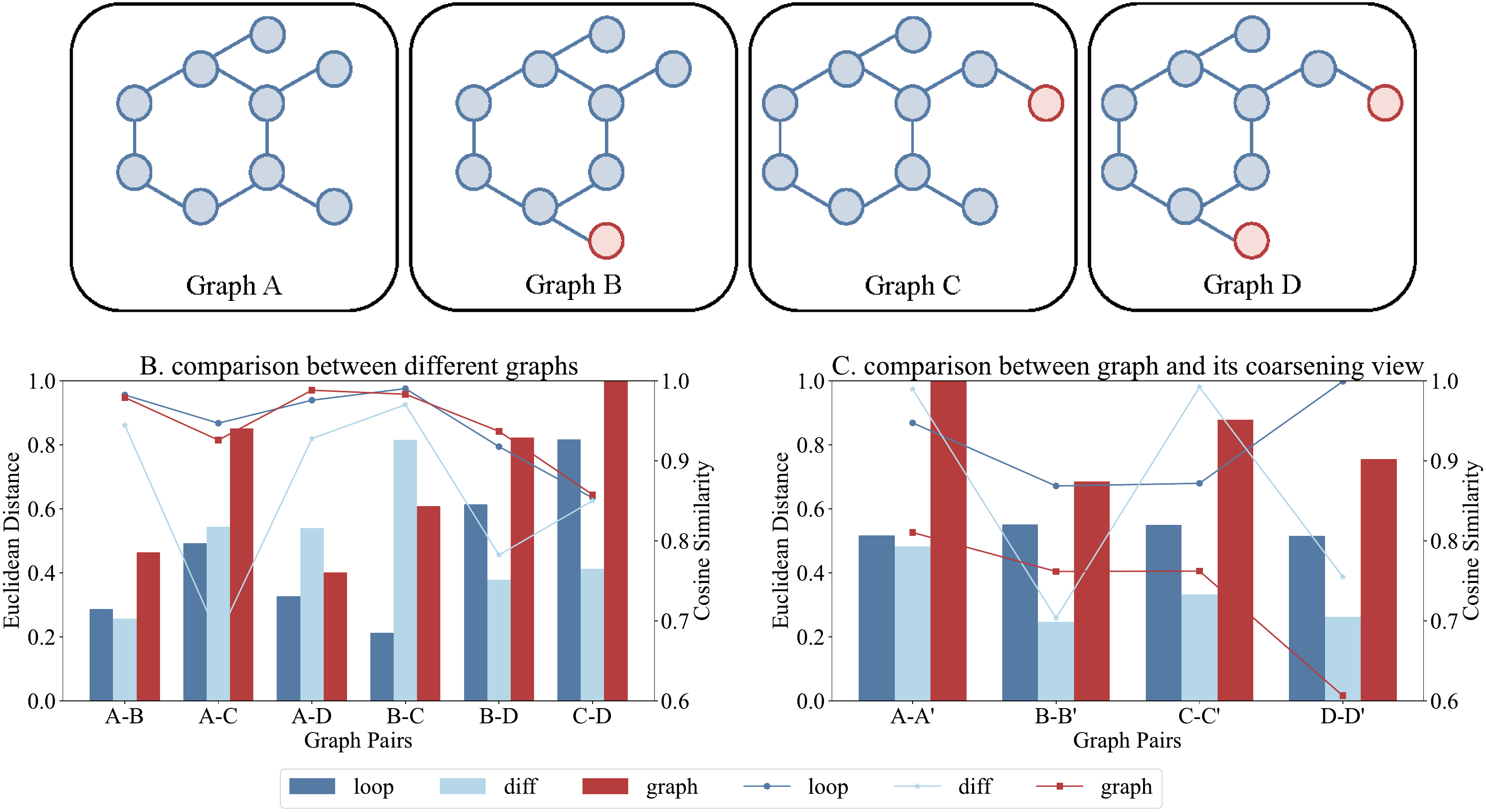}
    \caption{Pre-experiment on a toy molecule graph set, which shows four graphs with different features, all containing a benzene ring. \textbf{(B)} illustrates the pairwise comparison between four graphs. \textbf{(C)} presents the comparison between the graph and its coarsening view. We leverage two measurements of Euclidean Distance (bar plots, lower for closer) and Cosine Similarity (line plots, higher for closer) to reveal the latent relations between graphs.}
    \label{fig:pre}
\end{figure*}

We further explore the feasibility of representing structures via hypernodes through a set of pre-experiments.
Figure~\ref{fig:pre} shows the results.
We train a plain GCN and use summation as the readout function, obtaining three representations: the benzene ring (\emph{loop}), main distinguishing atom sets (\emph{diff}), and whole graph (\emph{graph}).
We measure the spatial distance and angular separation between representations by Euclidean Distance and Cosine Similarity, respectively.
Figure~\ref{fig:pre}(B) shows the comparison between different graphs.
The experiments present instructive results, where the gap between \emph{loop} is less, w.r.t. \emph{graph} and \emph{diff}.
It highlights that the prime factor causing the differences lies behind the main distinguish atom sets rather than the benzene ring.
Thus, we coarsen the benzene ring into one node with the same settings and obtain the coarsening view of each graph.
Figure~\ref{fig:pre}(C) shows the distance between the original graph and its coarsening view.
Pairs such as A and C, whose main distinguish atom sets are arranged adjacent, leave a relatively closer cosine similarity, while B and D show a significant difference.
It indicates that simple coarsening will lose part of structural information~\cite{ref:gc_survey}.
In summary, pre-experiments tell us two enlightenments:
(1) some structures contribute relatively less to distinguish graphs, which can further turn into a coarsening view to magnify the high-level structural information;
(2) after coarsening these structures into hypernodes, some of the structural information may be traceless, calling for additional consideration for the relative position of neighborhoods.

Some researchers have focused on this problem and defined the overall treatment idea as Graph Compression~\cite{ref:CoGSL}, Pooling~\cite{ref:PAS,ref:GMT,ref:MSGNN}, and Coarsening~\cite{ref:CGKS,ref:AI24}.
This series of works aims to compress the set of nodes into a compact representation~\cite{ref:SEPG} or emphasize the hierarchical structures of graphs~\cite{ref:DIFFPOOL}.
Despite all achievements, there are still many challenges.
\textbf{(1) Unsuitability coarsening.}
Some works aim to achieve better results on different tasks, such as graph reconstruction~\cite{ref:LV18} and node classification~\cite{ref:GCond}, where the coarsened perspective is not necessarily helpful for graph classification.
\textbf{(2) Poor efficiency.}
Identifying the whole set of structures (e.g., motifs~\cite{ref:motif}) has an unbearable computational overhead, especially for those condensing methods~\cite{ref:GCond,ref:DosCond}.
\textbf{(3) Distribution shift.}
Previous studies have shown that the model will exhibit poor generalization when there is a difference between the training and testing set distribution~\cite{ref:ds}.
A deeper coarsening layer reduces abundant graph elements, lacking an effective alignment of the original graph, which has poor interpretability~\cite{ref:gc_survey}.
It further hinders the credibility of the coarsening graph view.

Therefore, we propose an efficient coarsening algorithm based on the above findings and challenges.
Our model relies on the GT framework and aims to achieve better performance on graph classification tasks (challenge 1).
To address the issue posed by challenge 2, we focus on coarsening two widespread classes of structures, loops, and cliques as a trade-off and implement a linear algorithm with well-designed constraints.
Finally, we solve challenge 3 by introducing line graph conversion to construct a multi-view learning framework.
The multi-view fusion can preserve the original graph distribution for learning.
On the other hand, a shallow coarsening layer reveals clearer correspondences and provides more reliable interpretation guarantees.

The contributions of this paper are as follows:
\begin{itemize}
    \item We propose an efficient \textbf{L}oop and \textbf{C}lique \textbf{C}oarsening algorithm with linear complexity \textbf{for} \textbf{G}raph \textbf{C}lassification (LCC4GC), which captures high-level topological structures.
    \item We introduce line graph conversion to conduct multi-view graph representation learning on the GT framework, which learns more comprehensive and reliable structural information.
    \item We verify the performance of LCC4GC on graph classification tasks, compared with 28 baselines from six categories under eight datasets from three categories. LCC4GC achieves better results in most cases.
\end{itemize}

\section{Related Work}
\subsection{Graph Pooling}
Graph Pooling~\cite{ref:PAS,ref:GMT,ref:MSGNN} is one of the pivotal techniques for graph representation learning, aiming to compress the input graph into a smaller one.
Due to the hierarchy characteristics of graphs, hierarchical pooling methods have become mainstream nowadays.
Methods fall into two categories depending on whether breaking nodes.
The first category conducts new nodes via graph convolution.
The pioneer in hierarchical pooling, DIFFPOOL~\cite{ref:DIFFPOOL}, aggregates nodes layer by layer to learn high-level representations.
Inspired by the Computer Vision field, CGKS~\cite{ref:CGKS} constructs the pooling pyramid under the contrastive learning framework, extending the Laplacian Eignmaps with negative sampling.
MPool~\cite{ref:MPool} takes advantage of motifs to capture high-level structures based on motif adjacency.
The second category condenses graphs by sampling from old nodes.
SAGPool~\cite{ref:SAGPool} masks nodes based on top-rank selection to obtain the attention mask subgraphs.
CoGSL~\cite{ref:CoGSL} extracts views by perturbation, reducing the mutual information for compression.
MVPool~\cite{ref:MVPool} reranks nodes across different views with the collaboration of attention mechanism, then selects particular node sets to preserve underlying topological.
\subsection{Graph Coarsening and Condensation}
Similar to Graph Pooling, both Graph Coarsening~\cite{ref:CGIPool,ref:KGC,ref:SS} and Graph Condensation~\cite{ref:GCond,ref:DosCond} are graph reduction methods that simplify graphs while preserving essential characteristics~\cite{ref:gc_survey}.
Generally speaking, Graph Coarsening groups and clusters nodes into super node sets using specified aggregation algorithms.
It preserves specific graph properties considered high-level structural information, revealing hierarchical views of the original graph.
L19~\cite{ref:L19} proposes a restricted spectral approximation with relative eigenvalue error.
SS~\cite{ref:SS} conducts both intra- and inter-cluster features to strengthen the fractal structures discrimination power of models.
In addition to the above spectral methods, researchers try to find other ways to make measurements.
KGC~\cite{ref:KGC} calculates graphs equipped with Gromov-Wasserstein distance.

Graph Condensation, first introduced in~\cite{ref:GCond}, leverages learnable downstream tasks' information to minimize the loss between the original graph and some synthetic graphs.
They propose a method based on gradient matching techniques, called GCond~\cite{ref:GCond}, to condense the structures via MLP.
However, GCond involves a nested loop in steps.
Additional requirements for scalability led to the creation of DosCond~\cite{ref:DosCond}, EXGC~\cite{ref:EXGC}, and many other works~\cite{ref:gc_2023,ref:gc_2024}.
DosCond~\cite{ref:DosCond} is the first work focusing on graph classification tasks via graph condensation.
EXGC~\cite{ref:EXGC}, on the other hand, heuristically identifies two problems affecting graph condensation based on gradient matching and proposed solutions.
Both types of work hope to achieve the same downstream task effect as the original graph while reducing the graph scale.
For a further detailed understanding of those techniques, we recommend reading the survey~\cite{ref:gc_survey}.
\subsection{Graph Transformers}
Graph Transformers (GTs)~\cite{ref:KDD22,ref:GPTrans,ref:KDLGT} alleviate the problems of over-smoothing and local limitations of GNNs, which has attracted the attention of researchers.
The self-attention mechanism learns long-distance interactions between every node pair and shows tremendous potential for various scenarios.
Graphormer~\cite{ref:Graphormer} integrates edges and structures into Transformers as biases, outperforming in graph-level prediction tasks.
U2GNN~\cite{ref:U2GNN} alternates aggregation function leveraging Transformer architecture.
Exphormer~\cite{ref:Exphormer} builds an expansive GT leveraging a sparse attention mechanism with virtual node generation and graph expansion.
In addition, due to the quadratic computational constraint $O(N^2)$ of self-attention, several works~\cite{ref:NodeFormer,ref:Gapformer} have been proposed to focus on a scalable and efficient Transformer.
The former puts forward a new propagation strategy adapting arbitrary nodes in scale.
The latter combines pooling blocks before multi-head attention to shrink the size of fully connected layers. 
To summarize, we design our model based on the capabilities of GTs, which can effectively aggregate features of distant nodes using attention mechanisms and overcome the limitations of local neighborhoods.

\section{Preliminaries}
\paragraph{Graphs}
Given a graph $G = \{\mathcal{V}, \mathcal{E}\}$ where $\mathcal{V} = \{v_i\}_{i=1}^N$ and $\mathcal{E} \subseteq \mathcal{V} \times \mathcal{V}$ denote the set of nodes and edges, respectively.
We leverage $A \in \{0, 1\}^{N \times N}$ to indicate the adjacency matrix, and $A[i, j] = 1$ when there is an edge between node $v_i$ and $v_j$.
We use $H = \{h_{ij}\}^{N \times D}$ to denote the node features, where $D$ shows the dimension of feature space, and $h_{ij}$ represents the feature of node $v_i$ in dimension $j$.
\paragraph{Problem Definition} 
For supervised graph representation learning, given a set of graphs $\mathcal{G} = \{G_1, \cdots, G_M\}$ and their labels $\mathcal{Y} = \{y_1, \cdots, y_M\}$, our goal is to learn a representation vector $H_i$ for each graph, which can be used in downstream classification tasks to predict the label correctly. 
\paragraph{Universal Graph Transformer (U2GNN)}
U2GNN~\cite{ref:U2GNN} is one of the SOTAs in graph classification tasks, a GNN-based model that follows the essential aggregation and readout pooling functions.
Xu et al.~\cite{ref:GIN} claim that a well-designed aggregation function can further improve the performance.
Thus, U2GNN replaces the plain strategy which GNN uses with the following aggregation function:
\begin{equation}
    h_{t, v}^{(k)} = \mathbf{AGG}_{\mathsf{att}}(h_{t-1,v}^{(k)}),
    \label{eq:U2GNN}
\end{equation}
where $h_{t, v}^{(k)}$ denotes the representation vector of node $v$ in step $t$ of layer $k$, which aggregates from step $t-1$.
It provides a powerful method based on the transformer self-attention architecture to learn graph representation.
We take U2GNN as the backbone and cover the details later in Section 4.4.

\section{Methodology}
\subsection{Overview}
\begin{figure*}[!t]
    \centering
    \includegraphics[width=\linewidth]{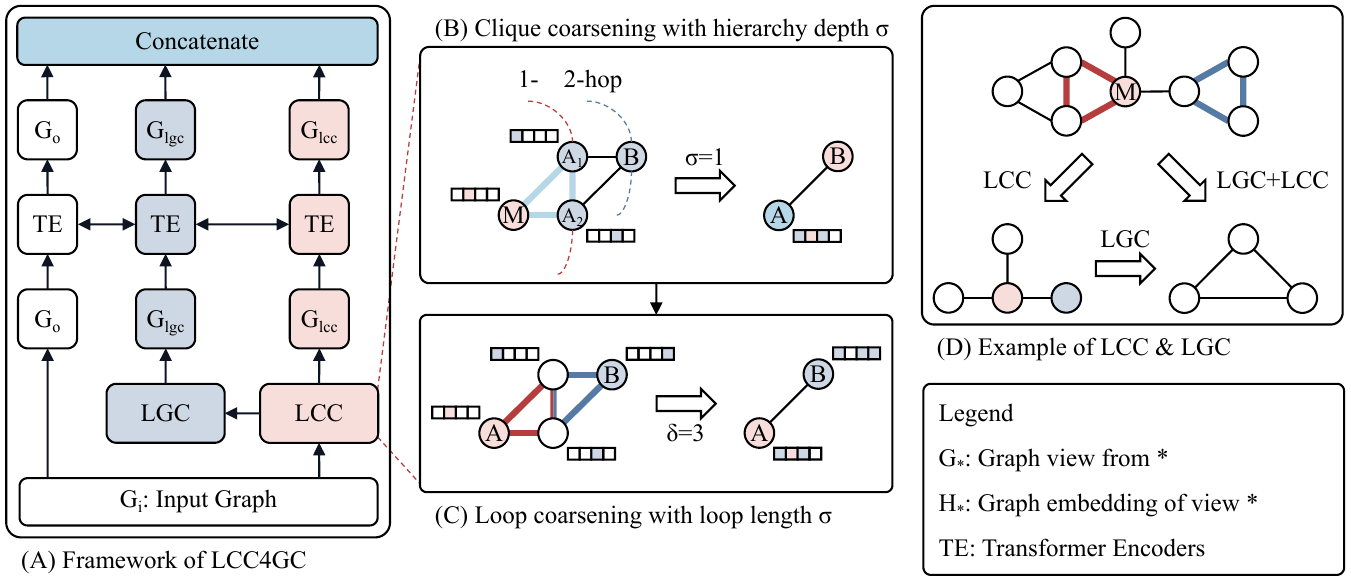}
    \caption{Overview of our work. (A) The framework of LCC4GC. LCC contains two procedures: (B) clique coarsening and (C) loop coarsening. We design two constraints, hierarchy depth, and loop length, to control the graph coarsening process. (D) An example of how LCC and LGC work.}
    \label{fig:frame}
\end{figure*}
Figure~\ref{fig:frame}(A) shows the framework diagram of our proposed model LCC4GC, which mainly contains three parts: multi-view construction of graphs, graph encoders, and classifier (omitted in the figure).
Given an input graph $G_i$, LCC4GC constructs three views step by step.
Keep the original graph as $G_o$, and coarsen $G_i$ via the \textbf{L}oop and \textbf{C}lique \textbf{C}oarsening (LCC) to build view $G_{lcc}$.
After that, we design the \textbf{L}ine \textbf{G}raph \textbf{C}onversion (LGC) to transfer $G_{lcc}$ to view $G_{lgc}$.
Subsequently, LCC4GC trains a GT for each view separately to obtain graph-level representation, which is directly concatenated and fed into the downstream classifier.

This chapter is structured as follows: 
Section 4.2 and Section 4.3 describe LCC and LGC, respectively. 
Section 4.4 explains the rest of the details of LCC4GC.
\subsection{Loop and Clique Coarsening}
\subsubsection{Algorithm Description}
As for coarsening structures into hypernodes, existing methods mainly pay attention to the hierarchical structure of graphs: some cluster nodes based on graph convolution to do graph coarsening~\cite{ref:DIFFPOOL}, and some implement node condensing by iterative sampling~\cite{ref:SAGPool}.
Aggregation of first-order adjacent nodes has intuitive interpretability, which means connections between neighborhoods, while a deeper coarsening shows a poor explanation.
Moreover, as mentioned in challenge 3, the particular node disappears into hypernodes, which leads to an ambiguity for topology.
Thus, LCC4GC achieves graph coarsening of loops and cliques with shallow coarse-grained clustering restricted by two hard constraints, which magnifies high-level structural information while preserving the characteristic nodes in graphs to the greatest extent.
We first give formal definitions of loops and cliques.
\begin{definition}[Loop]
    Given a graph $G$, a path $G_l = \{v_1\cdots v_{n+1}\}$ is a loop if there is no repetitive node in $\{v_1, \cdots, v_n, v_{n+1}\}$ except $v_1 = v_{n+1}$ and $(v_k, v_{k+1}) \in \mathcal{E}_{G}$ for $k = 1, 2, \cdots, n$.
    $G_l$ is a k-loop if $|\mathcal{V}_{G_l}| = k$.
\end{definition}
\begin{definition}[Clique]
    A subgraph $G_c$ is a clique if $\forall v_i, v_j \in \mathcal{V}_{G_c}, (v_i, v_j) \in \mathcal{E}_{G} (i \neq j)$.
    $G_c$ is a k-clique if $|\mathcal{V}_{G_c}|=k$.
\end{definition}

Coarsening requires counting all loops and cliques in the graph.
However, according to the definitions above, loops are contained within cliques.
For example, given a 4-clique marked $Cli = \{ABCD\}$, we can find four 3-loops: $\{ABC\}$, $\{ABD\}$, $\{ACD\}$, and $\{BCD\}$, and one 4-loop $\{ABCD\}$, except for one 4-clique, which is not we want, since identifying $Cli$ as a clique is much more straightforward than five loops.
Moreover, since the Maximum Clique Problem (MCP) is NP-hard~\cite{ref:MCP}, finding all cliques is also NP-hard.
Algorithms require an approximation or a shortcut due to enormous search space.

Mathematically, given an original graph $G_o = \{\mathcal{V}, \mathcal{E}\}$, LCC aims to rebuild an intermediate coarsening graph $G_{lcc}$ with $n$ nodes where $n \ll |\mathcal{V}|$.
In methods~\cite{ref:LV18,ref:L19}, supernode $v_i$ in $G_{lcc}$ aggregates nodes in $G_o$ according to node partitioning sets $\mathcal{P} = \{\mathcal{P}_1, \cdots, \mathcal{P}_n\}$.
We use indication matrix $M_p \in \{0, 1\}^{n \times |\mathcal{V}|}$ to denote the aggregation based on partition set $\mathcal{P}$, where $m_{ij}=1$ if node $j$ in partition $\mathcal{P}_i$.
We use a new adjacency matrix to represent $G_{lcc}$: 
\begin{equation}
    A_{lcc} = M_pAM_p^T,
    \label{eq:adjacency}
\end{equation}
where we multiply $M_p$ to the right to ensure the construction of the undirected graph.
Here, we notice that $DEG=diag(A_{lcc})$ gives weight $a_{ii}$ for each supernode $i$, representing the sum of all node degrees corresponding to the partition $\mathcal{P}_i$.
To directly extract and learn high-level structural information, we consider making a separation.
We assign the weight of supernodes to 1 with $DEG^{-1}=diag(a_{ii}^{-1}, \cdots, a_{nn}^{-1})$ instead of summation.
The normalized adjacency matrix is 
\begin{equation}
    \bar{A}_{lcc} = diag(a_{ii}^{-1}, \cdots, a_{nn}^{-1})(M_pAM_p^T),
    \label{eq:norm_adjacency}    
\end{equation}
leaving each supernode equal contribution to the high-level structure initially.
Similarly, we can define the node features of $G_{lcc}$ using $M_p$:
\begin{equation}
    H_{lcc} = M_pH, 
    \label{eq:fea}
\end{equation}
where $H_{*}$ indicates the feature representation of $*$. 
This equation is equivalent to summing the nodes according to the partition sets.

So far, the focus of graph coarsening falls on how to obtain the partitioning set $\mathcal{P}$.
LCC finds partitioning sets by searching for loops and cliques, where each $\mathcal{P}_i$ represents a loop or a clique.
In other words, the process of LCC is to find a linear transformation matrix $A_L$ about $A$, such that 
\begin{equation}
    A' = A_LAA_L^T = \begin{bmatrix}
        \mathcal{P}_1 & E_{12} & \cdots & E_{1n}\\
        E_{21} & \mathcal{P}_2 & \ddots & \vdots\\
        \vdots & \ddots & \ddots & \vdots \\
        E_{n1} & \cdots & \cdots & \mathcal{P}_n
    \end{bmatrix},
    \label{eq:linear}
\end{equation}
where the diagonal $\mathcal{P}_i = \{1\}^{k_i \times k_i}$ is the partitioning set containing $k_i$ nodes, and the non-diagonal $E_{ij} \in \{0, 1\}^{k_i \times k_j}$ records the connection between partition $\mathcal{P}_i$ and partition $\mathcal{P}_j$.
Next, by mapping each submatrix to 0 or 1 with the indicator function $\mathcal{I}(X)$, we can get the same result as Equation~\ref{eq:adjacency}, i.e,
\begin{equation}
    A_{lcc} = \mathcal{I}(A'),
    \label{eq:eq}
\end{equation}
\begin{equation}
    \mathcal{I}(X) = \begin{cases}
    k_i & X = \mathcal{P}_i \\
    1 & \exists x \in X, X = E_{ij}, x = 1 \\
    0 & \text{otherwise}
    \end{cases}.
    \label{eq:indicator}
\end{equation}

Finally, we give the pseudocode of how the LCC works, as shown in Algorithm~\ref{alg:alg}.
Based on Tarjan~\cite{ref:Tarjan} and Clique Percolation (CP)~\cite{ref:CP}, LCC heuristically iterates the graph hierarchy using loop length and hierarchy depth as constraints for pruning.
LCC finds cliques under depths less than $\sigma$ (lines 3-13) first, counting loops only when finding fewer or no cliques (lines 14-21).
When updating graphs, LCC reconstructs graphs to build the coarsening view.
Figures~\ref{fig:frame}(B) and (C) further depict the workflow of the LCC.
We go into details one by one.
\begin{algorithm}[!ht]
    \caption{\textbf{L}oop and \textbf{C}lique \textbf{C}oarsening (LCC)}
    \label{alg:alg}
    \begin{algorithmic}[1] 
        \REQUIRE the input graph $G_i$, the max loop length $\delta$, the max hierarchy depth $\sigma$;
        \ENSURE the coarsening view $G_{lcc}$;
        \STATE $G_{lcc} \xleftarrow{} G_i, G_{cc} \xleftarrow{} \emptyset, vis \xleftarrow{} \emptyset, V_{lcc} \xleftarrow{} V_i$;
        \STATE sort $V_{lcc}$ from high degrees to low;
        \WHILE{not all nodes in $vis$}
            \STATE $cur \xleftarrow{} V_{lcc}[0]$;
            \STATE fetch a clique $Cli$ not visited containing node $cur$ ;
            \STATE $G_{cc} \xleftarrow{update} Cli, vis \xleftarrow{} vis \cup Cli \setminus \{cur\}$;
            \IF{exist node set $rest$ in the neighborhoods of $cur$ but has not been visited yet}
                \STATE find cliques $Cli_{rest}$ in $\sigma$ depth according to $rest$;
                \STATE $G_{cc} \xleftarrow{update} Cli_{rest}, vis \xleftarrow{} vis \cup Cli_{rest} \setminus \{cur\}$;
            \ELSE
                \STATE $vis \xleftarrow{} vis \cup cur$;
            \ENDIF
        \ENDWHILE
        \IF {less or no cliques found}
            \STATE $G_{lc} \xleftarrow{} \emptyset$;
            \FORALL{$v_i \in G_{lcc}$}
                \STATE find all loops $L_i$ in $\delta$ length beginning with $v_i$;
                \STATE $G_{lc} \xleftarrow{update} L_i$;
            \ENDFOR
            \STATE $G_{lcc} \xleftarrow{update} \{G_{cc}, G_{lc}\}$;
        \ENDIF
        \STATE \textbf{return} $G_{lcc}$
    \end{algorithmic}
\end{algorithm}
\subsubsection{Clique Coarsening with Hierarchy Depth Constraint.}
The NP-hard nature of finding all cliques determines the impossibility of exhaustive search, especially for large graphs.
LCC4GC takes advantage of the hierarchical characteristic of graphs, taking nodes with the highest degree (Alg~\ref{alg:alg} line 2) hop by hop to find cliques.
We set a distance $\sigma$ to control the depth of the recursion.
Figure~\ref{fig:frame}(B) illustrates the clique coarsening process with hierarchy depth.
Starting from central node $M$ with the highest degree, LCC searches for possible cliques hop by hop until $\sigma$ limits.
Cliques exceeding the limitation will not be considered under the current node neighbor.
For example, since $B$ is 2-hop far from $M$, which exceeds the depth constraint $\sigma=1$, LCC only coarsens clique $\{MA_1A_2\}$ and leaves $\{BA_1A_2\}$ alone, where $\mathcal{P} = \{\{MA_1A_2\}, \{B\}\}$.
When $M$ and its neighbors are processed, LCC switches the central node to $B$ and conducts a new $\sigma$-hop coarsening.
\subsubsection{Loop Coarsening with Loop Length Constraint.}
Not all loops are suitable for coarsening.
Long loops may reveal two chains or sequences interacting (e.g., backbone in proteins~\cite{ref:3D}).
The semantics will change if coarsening happens.
LCC4GC sets a range $\delta$ for the maximum detection length (Alg ~\ref{alg:alg} line 17).
We default it to 6 to cover usual cases, such as squares and benzene rings.
When the sequence length exceeds the constraint, we can make the pruning on that chain.
For example, in Figure~\ref{fig:frame}(C), there are two 3-loop marked by node $A$ in red and $B$ in blue.
The entire graph also includes a 4-loop.
When $\delta=3$, only loops under length limitation will be selected, where $\mathcal{P} = \{\{A, ...\}, \{B, ...\}\}$.
Note that 3-loop is equal to 3-clique.
We can handle it either way.
\subsubsection{Time Complexity}
According to the previous description, LCC contains two parts, clique and loop coarsening, and the total time complexity of the algorithm is the sum of them.
Given the input graph $G=\{\mathcal{V}, \mathcal{E}\}$, we search for cliques following the hierarchy depth $\sigma$, marking nodes visited if no possible neighbors can form the new cliques.
The time complexity of clique coarsening is $\mathcal{O}(\mathcal{V}+\mathcal{E})$.
$\sigma$ only gives an order for searching and does not influence the linear computational cost.
As for loop coarsening, we traverse along the DFS sequence and only keep loops under length constraints $\delta$.
The time complexity of loop coarsening is $\mathcal{O}(\delta \mathcal{V}+\mathcal{E})$.
For each node, we look back within $\delta$-level seeking loops satisfying the needs.
Since $\delta$ is relatively small, the computation stays linear, $\mathcal{O}(\mathcal{V}+\mathcal{E})$.
Therefore, the time complexity of the final LCC is $\mathcal{O}(2(\mathcal{V}+\mathcal{E}))$, achieving a linear growth in the scale of the input graph.
\subsection{Line Graph Conversion}
\begin{figure*}[!t]
    \centering
    \includegraphics[width=\linewidth]{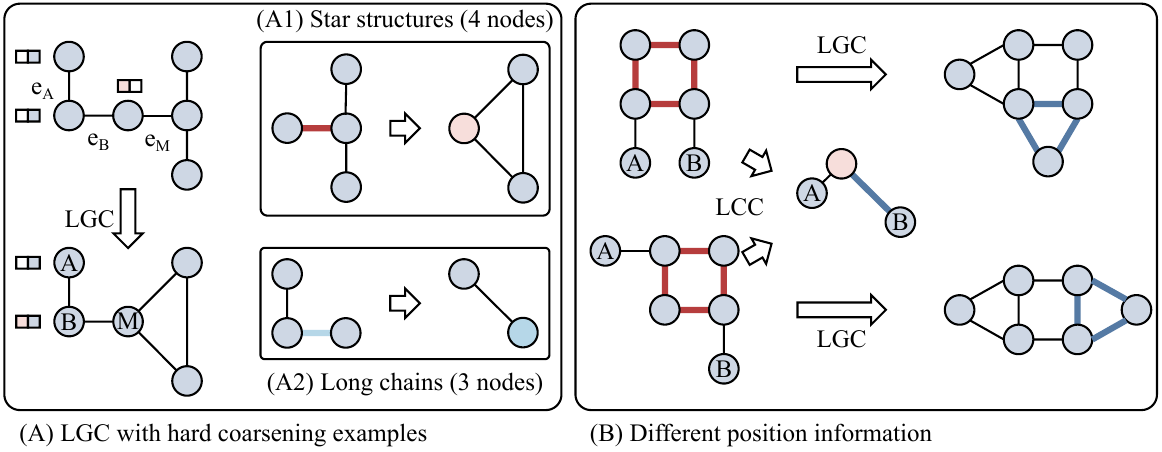}
    \caption{LGC workflow. (A) shows the line graph conversion with hard coarsening examples. (B) illustrates how the input graphs with different position information are distinguished by LGC but not LCC.}
    \label{fig:LGC}
\end{figure*}
Edge augmentation is a considerable view to make a thorough representation integrating various architectures and scenarios~\cite{ref:EPT,ref:Gapformer}.
Therefore, we attach an edge-central perspective through LGC to enrich the representation and alleviate the impact of coarsening reduction.
Firstly, the definition of line graphs is as follows.

\begin{definition}[Line Graph]
    Given a graph $G$, $L(G) = \{L(\mathcal{E}), \mathcal{E}_{L(G)}\}$ is the line graph of $G$, where $L(\mathcal{E})$ and $\mathcal{E}_{L(G)}$ denote the node set and edge set of the line graph, and $L$ is a mapping function from the edge set of the original graph to the node set of the line graph, satisfying $\forall e_{ij} \in \mathcal{E}, L(e_{ij}) \in L(\mathcal{E})$, and $\forall e_{ij}, e_{ik} \in \mathcal{E}, (L(e_{ij}), L(e_{ik})) \in \mathcal{E}_{L(G)} (j \neq k)$.
\end{definition}

In plain words, a line graph turns the edges in the original graph into new vertices and the endpoints of two edges in the original graph into new edges.
This explicit transformation can reduce the difficulty of relative position modeling, which is implicit in the node-central view of the original graph.
For those graphs with no edge attributes, we sum features of endpoints for consistency following Equation~\ref{eq:fea}.
In other words, we takes edges as the partitioning set and further aggregates features to build the LGC view.
Mathematically, given an input graph $G_o$ and its feature vector $H$, LGC computes: 
\begin{equation}
    H_{lgc} = M_e[L(H_{lcc})] = M_e[L(M_pH)],
    \label{eq:Hlgc}
\end{equation}
where $L(\cdot)$ is the conversion function according to definition 4.3, $M_e$ is the indication matrix with edge partitions, and $H_{lcc}$ is the feature representations output by Equation~\ref{eq:fea}.

In previous sections, we know LCC is limited if no loops and cliques exist, and the model suffers relative position information loss if we coarsen structures into hypernodes.
LGC is a supplementary of the coarsening view.
We explain this from two aspects: some hard-coarsening examples and different position information of neighborhood nodes.
Figure~\ref{fig:LGC} shows the details.
\subsubsection{LGC with Hard Coarsening Examples}
LCC4GC only focuses on loops and cliques.
However, not all graphs contain either structures, such as long chains.
A simple copy view contributes less if we leave these hard-coarsening examples alone.
LGC transforms these graphs into a fittable form that is easy for postprocessing while preserving the original structural information.
For example, Figures~\ref{fig:LGC}(A) and (B) give two hard-coarsening examples, neither of which belongs to loops and cliques.
LGC is sensitive to them, while LCC fails to handle them efficiently.
The star structures (Figure~\ref{fig:LGC}(A1)), all edges connected through the central node, form a new clique after conversion.
The long chain (Figure~\ref{fig:LGC}(A2)) with $n$ nodes can decline the length to $n-1$ in one LGC calculation, reducing the scale effectively.
LCC4GC applies one LGC step on the intermediate coarsening graph of LCC, retaining the newly generated structures after conversion.
\subsubsection{Different Position Information}
From pre-experiments, we know that relative position suffers if we condense graph structures into one node.
Though plain coarsening strategies decrease the graph scale, it makes positions vague, such as between Graphs A and B in Figure~\ref{fig:pre}.
In essence, position information is the relative position among nodes connected by node-edge sequence.
LGC models edge-central positional relationships, describing structural information by adding a new view to obtain a more comprehensive representation.
Figure~\ref{fig:LGC}(B) demonstrates this dilemma.
The two sample graphs differ only in the connection positions of nodes $A$ and $B$, one at adjacent nodes in the square and the other at diagonal nodes.
Both point to the same coarsening view after LCC, which fails to distinguish.
LGC solves this problem by expanding the edge information of node $B$, further showing a conspicuous difference of cliques in blue.
\subsection{LCC4GC: Model Architecture}
After acquiring three views, we can build our model LCC4GC based on U2GNN~\cite{ref:U2GNN}.
As shown in Figure~\ref{fig:frame}, LCC4GC takes $G_i$ as input and obtains the coarsening view $G_{lcc}$ and line graph conversion view $G_{lgc}$ with LCC and LGC, respectively.
For $G_{lcc}=\{\mathcal{V}_{lcc}, \mathcal{E}_{lcc}\}$, LCC4GC calculates node embeddings $h$ according to Equation~\ref{eq:U2GNN} and Equation~\ref{eq:fea} and uses the summation readout function to obtain graph embedding $H_{lcc}$.
Specifically, it is
\begin{equation}
    H_{lcc} = \sum_{v \in \mathcal{V}_{lcc}} [h_{0, v}^1; \cdots; h_{0, v}^K],
\end{equation}
where $h_{0, v}^k$ denotes the initial embedding of node $v$ in layer $k$.
For each layer $k$, LCC4GC iterates $T$ steps aggregating $N$ sampling neighbors and passes $h_{T,v}^k$ to the next layer $k+1$.

We use the same operation for the other two views, $G_o$ and $G_{lgc}$, and achieve corresponding embeddings, $H_o$ and $H_{lgc}$.
We apply concatenation across views to obtain the final embedding $H_i$ for the input graph $G_i$ as
\begin{equation}
    H_i = [H_o; H_{lcc}; H_{lgc}].
\end{equation}

After that, we feed the embedding $H_i$ to a single fully connected (FC) layer.
The loss function is cross-entropy as follows:
\begin{equation}
    loss = -\sum_{i=1}^{M} y_i \log(\sigma(\hat{y}_i)),
    \label{eq:loss}
\end{equation}
where $\sigma$ denotes the softmax function.
For further implementation details, we suggest to refer to ~\cite{ref:U2GNN}.

\section{Experiments}
\begin{table}[!t]
    \centering
    \caption{The statistics of datasets from TUDataset. Avg.N, Avg.E, and Avg.D denote the average number of nodes, edges, and degrees. SN for social network, BIO for bioinformatics, and MOL for molecules.}
    \begin{tabular}{cccrrrc}
        \hline
        \bf{Dataset} & \bf{\# G} & \bf{\# Cl} & \bf{Avg.N} & \bf{Avg.E} & \bf{Avg.D} & \bf{Category} \\
        \hline
        CO & 5000 & 3 & 74.49 & 2457.78 & 37.39 & SN \\
        DD & 1178 & 2 & 284.32 & 715.66 & 4.98 & BIO \\
        PTC & 344 & 2 & 25.56 & 25.96 & 1.99 & BIO \\
        PRO & 1113 & 2 & 39.06 & 72.82 & 3.73 & BIO \\
        IB & 1000 & 2 & 19.77 & 96.53 & 8.89 & SN \\
        IM & 1500 & 3 & 13.00 & 65.94 & 8.10 & SN \\
        N1 & 4110 & 2 & 29.87 & 32.30 & 2.16 & MOL \\
        N109 & 4127 & 2 & 29.68 & 32.13 & 2.16 & MOL \\
        \hline
    \end{tabular}
    \label{tab:TUDataset}
\end{table}
\subsection{Experimental Settings}
\subsubsection{Datasets}
We evaluate our approach on eight datasets from TUDataset~\cite{ref:TUDataset}, including three social network datasets: COLLAB (CO), IMDB-BINARY (IB), and IMDB-MULTI (IM), two molecules datasets: NCI1 (N1) and NCI109 (N109), and three bioinformatics datasets: D\&D (DD), PTC\_MR (PTC), and PROTEINS (PRO).
Table~\ref{tab:TUDataset} shows the details of the datasets.
\subsubsection{Baselines}
To fully evaluate the effectiveness of our model, we select 31 related works from 6 categories for comparison.

As for the Graph Transformer framework, we pick 20 baselines.
They are 
\textbf{(I) 2 kernel-based methods}: GK~\cite{ref:GK} and WL~\cite{ref:WL}; 
\textbf{(II) 8 GNN-based methods}: GCN~\cite{ref:GCN}, GAT~\cite{ref:GAT}, GraphSAGE~\cite{ref:GraphSAGE}, DGCNN~\cite{ref:DGCNN}, CapsGNN~\cite{ref:CapsGNN}, GIN~\cite{ref:GIN}, GDGIN~\cite{ref:GDGIN}, and GraphMAE~\cite{ref:GraphMAE}; 
\textbf{(III) 4 Contrastive Learning methods}: InfoGraph~\cite{ref:InfoGraph}, JOAO~\cite{ref:JOAO}, RGCL~\cite{ref:RGCL}, and CGKS~\cite{ref:CGKS};
and \textbf{(IV) 6 GT-based methods}: GKAT~\cite{ref:GKAT}, GraphGPS~\cite{ref:GraphGPS}, U2GNN~\cite{ref:U2GNN}, SAN-SSCA~\cite{ref:SAN-SSCA}, SA-GAT~\cite{ref:SA-GAT}, and UGT\cite{ref:UGT}.

We also choose 11 models from 2 categories to verify our Graph Coarsening technique. 
One is \textbf{(V) 7 Graph Pooling methods}: DIFFPOOL~\cite{ref:DIFFPOOL}, SAGPool\cite{ref:SAGPool}, ASAP~\cite{ref:ASAP}, GMT~\cite{ref:GMT}, SAEPool~\cite{ref:SAEPool}, MVPool~\cite{ref:MVPool}, and PAS~\cite{ref:TOISPAS}. 
The other is \textbf{(VI) 4 Graph Coarsening and Condensing methods}: CGIPool~\cite{ref:CGIPool}, DosCond~\cite{ref:DosCond}, KGC~\cite{ref:KGC}, and SS~\cite{ref:SS}.

To further explain the unique design of our LCC, we select 3 coarsening techniques: 
networkx~\cite{ref:networkx}, KGC~\cite{ref:KGC}, and L19~\cite{ref:L19}. 
Each one takes two variants: neighborhood and clique. 
We will explain the details later in Section 5.5.
\subsubsection{Implementation Details}
We follow the work~\cite{ref:GIN,ref:U2GNN} to evaluate the performance of our proposed method, adopting accuracy as the evaluation metric for graph classification tasks.
To ensure a fair comparison between methods, we use the same data splits and 10-fold cross-validation accuracy to report the performance.
All experiments are trained and evaluated on an NVIDIA RTX 3050 OEM 8GB GPU.
Our code is available: \url{https://github.com/NickSkyyy/LCC-for-GC}.
\subsection{Graph Classification Results}
\begin{table*}[!hpt]
    \caption{The main results of the graph classification task on four of eight datasets (mean accuracy (\%) and standard deviation). The best scores are in \underline{\bf{bold} and underline}. '-' indicates that results are unavailable in the original or published papers. A.R shows the average ranking of each method. We highlight every method with outstanding performance in each category.}
    \centering
    \begin{tabular}{lccccr}
        \hline
        \bf{Method} & \bf{CO} & \bf{DD} & \bf{PTC} & \bf{PRO} & \bf{A.R}\\
        \hline
        GW & $72.84^{0.28}$ & $78.45^{0.26}$ & $57.26^{1.41}$ & $71.67^{0.55}$ & 21.00\\
        WL & $79.02^{1.77}$ & $79.78^{0.36}$ & $57.97^{0.49}$ & $74.68^{0.49}$ & \underline{\textbf{9.63}}\\
        \hline
        GCN & $81.72^{1.64}$ & $79.12^{3.07}$ & $59.40^{10.30}$ & $75.65^{3.24}$ & 10.25\\
        GAT & $75.80^{1.60}$ & $74.40^{0.30}$ & $66.70^{5.10}$ & $74.70^{2.20}$ & 16.14\\
        GraphSAGE & $79.70^{1.70}$ & $65.80^{4.90}$ & $63.90^{7.70}$ & $65.90^{2.70}$ & 15.13\\
        DGCNN & $73.76^{0.49}$ & $79.37^{0.94}$ & $58.59^{2.47}$ & $75.54^{0.94}$ & 16.14\\
        CapsGNN & $79.62^{0.91}$ & $75.38^{4.17}$ & $66.00^{1.80}$ & $76.28^{3.63}$ & 10.71\\
        GIN & $80.20^{1.90}$ & $75.20^{2.90}$ & $64.60^{7.00}$ & $76.20^{2.80}$ & 8.88\\
        GDGIN & - & $77.80^{3.60}$ & $60.30^{4.50}$ & $73.70^{3.40}$ & 17.33\\
        GraphMAE & $80.32^{0.46}$ & - & - & $75.30^{0.39}$ & \underline{\textbf{6.60}}\\
        \hline
        InfoGraph & $70.65^{1.13}$ & $72.85^{1.78}$ & $61.65^{1.43}$ & $74.40^{0.30}$ & 17.43\\
        JOAO & $75.53^{0.18}$ & $75.81^{0.73}$ & - & $73.31^{0.48}$ & 19.50\\
        RGCL & $70.92^{0.65}$ & $78.86^{0.48}$ & - & $75.03^{0.43}$ & 17.25\\
        CGKS & $76.80^{0.10}$ & - & $63.50^{1.30}$ & $76.00^{0.20}$ & \underline{\textbf{10.00}}\\
        \hline
        DIFFPOOL & $75.48^{0.00}$ & $80.64^{0.00}$ & - & $76.25^{0.00}$ & 13.14\\
        SAGPool & $78.85^{0.56}$ & $76.45^{0.97}$ & - & $71.86^{0.97}$ & 16.14\\
        ASAP & $78.64^{0.50}$ & $76.87^{0.70}$ & - & $74.19^{0.79}$ & 15.14\\
        GMT & $80.74^{0.54}$ & $78.72^{0.59}$ & - & $75.09^{0.59}$ & 10.20\\
        SAEPool & - & - & - & $80.36^{0.00}$ & 9.00\\
        MVPool & - & $82.20^{1.40}$ & - & $\underline{\bf{85.70}}^{1.20}$ & 7.20\\
        PAS & - & $79.62^{1.75}$ & - & $77.36^{3.69}$ & \underline{\textbf{6.00}}\\
        \hline
        GKAT & $73.10^{2.00}$ & $78.60^{3.40}$ & - & $75.80^{3.80}$ & 16.50\\
        GraphGPS & - & - & - & $53.75^{6.20}$ & 13.67\\
        U2GNN & $77.84^{1.48}$ & $80.23^{1.48}$ & $69.63^{3.60}$ & $78.53^{4.07}$ & 4.67\\
        SAN-SSCA & $73.96^{0.00}$ & $82.41^{0.00}$ & - & - & 5.50\\
        SA-GAT & - & $91.40^{0.55}$ & $66.93^{1.49}$ & $77.80^{0.58}$ & \underline{\textbf{3.80}}\\
        UGT & - & - & - & $80.12^{0.32}$ & 7.67\\
        \hline
        CGIPool & $80.30^{0.69}$ & - & - & $74.10^{2.31}$ & 10.67\\
        DosCond & - & $78.92^{0.64}$ & - & - & 17.00\\
        KGC & - & - & $61.58^{2.49}$ & $66.43^{0.92}$ & 20.00\\
        SS & - & $79.78^{0.49}$ & - & $76.13^{0.60}$ & \underline{\textbf{7.25}}\\
        \hline
        LCC4GC & $\underline{\bf{98.18}}^{0.42}$ & $\underline{\bf{99.58}}^{0.78}$ & $\underline{\bf{76.43}}^{5.22}$ & $76.37^{4.37}$ & \underline{\textbf{3.25}}\\
        \hline
    \end{tabular}
    \label{tab:main_result_1}
\end{table*}
\begin{table*}[!ht]
    \caption{The rest four datasets due to limited space.}
    \centering
    \begin{tabular}{lccccr}
        \hline
        \bf{Method} & \bf{IB} & \bf{IM} & \bf{N1} & \bf{N109} & \bf{A.R}\\
        \hline
        GW & $65.87^{0.98}$ & $43.89^{0.38}$ & $62.30^{0.30}$ & - & 21.00\\
        WL & $73.40^{4.63}$ & $49.33^{4.75}$ & $82.19^{0.18}$ & $82.46^{0.24}$ & \underline{\textbf{9.63}}\\
        \hline
        GCN & $73.30^{5.29}$ & $51.20^{5.13}$ & $76.00^{0.90}$ & $67.09^{3.43}$ & 10.25\\
        GAT & $70.50^{2.30}$ & $47.80^{3.10}$ & $74.90^{0.10}$ & - & 16.14\\
        GraphSAGE & $72.40^{3.60}$ & $49.90^{5.00}$ & - & $64.67^{2.41}$ & 15.13\\
        DGCNN & $70.03^{0.86}$ & $47.83^{0.85}$ & $74.44^{0.47}$ & - & 16.14\\
        CapsGNN & $73.10^{4.83}$ & $50.27^{2.65}$ & $78.35^{1.55}$ & - & 10.71\\
        GIN & $75.10^{5.10}$ & $52.30^{2.80}$ & $79.10^{1.40}$ & $68.44^{1.89}$ & 8.88\\
        GraphMAE & $75.52^{0.66}$ & $51.63^{0.52}$ & $80.40^{0.30}$ & - & \underline{\textbf{6.60}}\\
        \hline
        InfoGraph & $73.03^{0.87}$ & $49.69^{0.53}$ & $73.80^{0.70}$ & - & 17.43\\
        JOAO & $70.21^{3.08}$ & $47.22^{0.41}$ & $74.86^{0.39}$ & - & 19.50\\
        RGCL & $71.85^{0.84}$ & - & - & - & 17.25\\
        CGKS & - & - & $79.10^{0.20}$ & - & \underline{\textbf{10.00}}\\
        \hline
        DIFFPOOL & $73.14^{0.70}$ & $49.53^{3.98}$ & $62.32^{1.90}$ & $61.98^{1.98}$ & 13.14\\
        SAGPool & $72.55^{1.28}$ & $49.33^{4.90}$ & $74.18^{1.20}$ & $74.06^{0.78}$ & 16.14\\
        ASAP & $72.81^{0.50}$ & $50.78^{0.75}$ & $71.48^{0.42}$ & $70.07^{0.55}$ & 15.14\\
        GMT & $73.48^{0.76}$ & $50.66^{0.82}$ & - & - & 10.20\\
        SAEPool & - & - & $74.48^{0.00}$ & $75.85^{0.00}$ & 9.00\\
        MVPool & $52.00^{0.80}$ & - & $80.10^{1.30}$ & $81.90^{1.60}$ & 7.20\\
        PAS & - & $53.13^{4.49}$ & - & - & \underline{\textbf{6.00}}\\
        \hline
        GKAT & $71.40^{2.60}$ & $47.50^{4.50}$ & $75.20^{2.40}$ & - & 16.50\\
        GraphGPS & - & - & $79.44^{0.65}$ & $76.27^{0.95}$ & 13.67\\
        U2GNN & $77.04^{3.45}$ & $53.60^{3.53}$ & - & - & 4.67\\
        SAN-SSCA & - & - & $\underline{\textbf{84.92}}^{0.00}$ & $\underline{\textbf{82.82}}^{0.00}$ & 5.50\\
        SA-GAT & $75.59^{0.77}$ & - & $79.32^{1.09}$ & - & \underline{\textbf{3.80}}\\
        UGT & - & - & $77.55^{0.16}$ & $75.45^{1.26}$ & 7.67\\
        \hline
        CGIPool & $72.40^{0.87}$ & $51.45^{0.65}$ & $78.62^{1.04}$ & $77.94^{1.37}$ & 10.67\\
        DosCond & - & - & $71.70^{0.20}$ & - & 17.00\\
        KGC & $69.20^{1.37}$ & - & - & - & 20.00\\
        SS & $73.89^{0.70}$ & $51.80^{1.20}$ & - & - & \underline{\textbf{7.25}}\\
        \hline
        LCC4GC & $\underline{\bf{89.40}}^{2.50}$ & $\underline{\bf{63.20}}^{3.55}$ & $79.08^{1.92}$ & $76.98^{1.90}$ & \underline{\textbf{3.25}}\\
        \hline
    \end{tabular}
    \label{tab:main_result_2}
\end{table*}
Tables~\ref{tab:main_result_1} and ~\ref{tab:main_result_2} show the main results of the graph classification task.
LCC4GC outperforms other baseline models on most datasets, with the highest average ranking 3.25 above all.
In detail, the boost of LCC4GC is higher in social networks than in molecules and bioinformatics.
WL, GraphMAE, CGKS, PAS, SA-GAT, and SS show competitive results among their categories, even better than LCC4GC (e.g., WL in N1 and N109).
The main reason is that cliques of social networks are more distinct, while plain cliques and loop structures can not cover functional molecules and proteins completely.
These main experimental results show the limitation of our model: the classification ability slightly drops when fewer loops and cliques appear.
N1 and N109 contain many simple loops and long chains, while some unconnected graphs in PRO make it hard to coarsen and convert.
We will show more details later in the ablation of the LCC  (Section 5.5) and case study (Section 5.6).
\subsection{Block Ablation Study}
\begin{table*}[!ht]
    \caption{Block ablation studies on all eight datasets.}
    \centering
    \begin{tabular}{lcccc}
        \hline
        \bf{Method} & \bf{CO} & \bf{DD} & \bf{PTC} & \bf{PRO} \\
        \hline
        LCC4GC & $\underline{\bf{98.18}}^{0.42}$ & $\underline{\bf{99.58}}^{0.78}$ & $\underline{\bf{76.43}}^{5.22}$ & $\underline{\bf{76.37}}^{4.37}$ \\
        $_{w/o \, LGC}$ & $97.96^{0.56}$ & $99.32^{1.06}$ & $69.50^{6.24}$ & $75.02^{4.12}$ \\
        $_{w/o \, LCC}$ & $97.60^{0.81}$ & $83.78^{4.63}$ & $66.92^{8.38}$ & $74.48^{3.42}$ \\
        $_{w/o \, both}$ & $96.50^{0.84}$ & $83.02^{5.33}$ & $63.09^{7.90}$ & $68.01^{3.13}$ \\
        \hline
    \end{tabular}
    \begin{tabular}{lcccc}
        \hline
        \bf{Method} & \bf{IB} & \bf{IM} & \bf{N1} & \bf{N109}\\
        \hline
        LCC4GC & $\underline{\bf{89.40}}^{2.50}$ & $\underline{\bf{63.20}}^{3.55}$ & $\underline{\bf{79.08}}^{1.92}$ & $\underline{\bf{76.98}}^{1.90}$ \\
        $_{w/o \, LGC}$ & $87.60^{2.58}$ & $61.47^{3.10}$ & $77.42^{1.64}$ & $74.90^{2.05}$ \\
        $_{w/o \, LCC}$ & $87.10^{2.91}$ & $58.60^{3.74}$ & $75.33^{2.12}$ & $72.79^{2.14}$ \\
        $_{w/o \, both}$ & $84.10^{2.91}$ & $57.87^{3.98}$ & $72.53^{2.28}$ & $69.20^{1.04}$ \\
        \hline
    \end{tabular}
    \label{tab:ablation}
\end{table*}
In this section, we will evaluate the effectiveness of each component of our model.
We design two sets of ablation experiments, one for modules to analyze the effectiveness of LCC4GC.
And the other to discuss the impact of different coarsening strategies on downstream classification tasks.
There are two variants of LCC4GC, LCC4GC$_{w/o \, LCC}$ removing the LCC and LCC4GC$_{w/o \, LGC}$ omitting the LGC.
Once we remove both components, LCC4GC degrades to U2GNN with a slight difference in the loss function and classifier.

As shown in Table~\ref{tab:ablation}, removing arbitrary components leads to performance degradation.
LCC4GC$_{w/o \, LGC}$ gets the most significant decrease by 6.93\% approximately on the PTC dataset, while LCC4GC$_{w/o \, LCC}$ gets 15.80\% on the DD dataset.
We observe that removing the LCC leads to a more significant decrease than LGC, which indicates that the global structural information introduced by the LCC is more vital for downstream classification tasks.
\subsection{Hyper-parameters Study}
\begin{figure*}[!t]
    \centering
    \includegraphics[width=\linewidth]{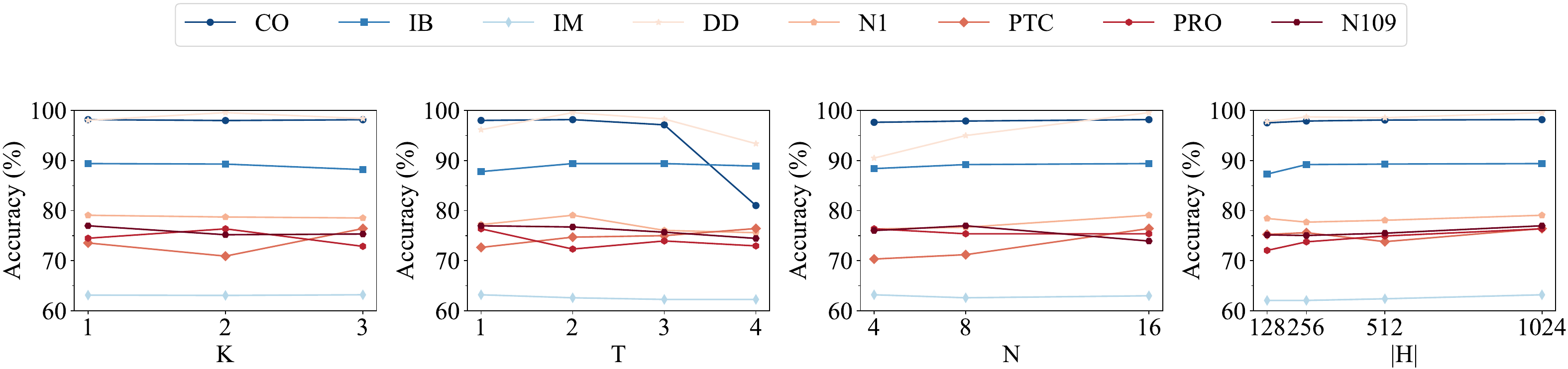}
    \caption{Hyper-parameters Study. We evaluate four hyper-parameters: hierarchy depth $K$, the number of Transformer Encoder layers $T$, the number of sampling neighbors $N$, and the hidden size $|H|$.}
    \label{fig:hyper}
\end{figure*}

In this section, we explore the hyper-parameters of our model.
We conduct experiments with different settings for 4 hyper-parameters: $K$, $T$, $N$, and hidden size on all eight datasets.
Table~\ref{tab:best_hyper} shows the optimal parameter settings, and Figure~\ref{fig:hyper} shows the results of hyper-parameters studies.
Generally, the optimal parameter settings are not the same for different datasets due to various data characteristics.
For a single hyper-parameter, we can observe some patterns in results.

As for $K$ and $T$, a deeper and more complicated model improves model performances and enhances the ability to capture complex structural information.
However, as the hierarchy extends continuously, overfitting leads to a decline in classification accuracy.
The most striking example is the plot of CO and DD on $T$, where the performance of LCC4GC decreases substantially when the number of layers changes from 3 to 4.

As for $N$ and $|H|$, in most cases, the larger the number of sampling neighbors and hidden size, the better the model performance.
It indicates that the increase in the sampling and hidden size within a particular range can promote the model to learn high-level structures thoroughly and clearly.
Among all datasets, IM, PRO, and N109 do not conform to the rules summarized above.
IM and PRO can capture long-distance structural information with the help of deeper hierarchy depth without sampling more neighbors.
N109 achieves a higher neighbor coverage with only a small amount of sampling for its multi-loop distribution (see Section 5.6 below for more details).
\begin{table*}[!ht]
    \centering
    \caption{Optimal parameter settings. We set $|H| = 1024$.}
    \begin{tabular}{ccccccccc}
        \hline
        \bf{Parameters} & \bf{CO} & \bf{DD} & \bf{PTC} & \bf{PRO} & \bf{IB} & \bf{IM} & \bf{N1} & \bf{N109}\\
        \hline
        $K$ & 1 & 2 & 3 & 2 & 1 & 3 & 1 & 1 \\
        $T$ & 2 & 2 & 4 & 1 & 2 & 1 & 2 & 1 \\
        $N$ & 16 & 16 & 16 & 4 & 16 & 4 & 16 & 8 \\
        \hline
    \end{tabular}
    \label{tab:best_hyper}
\end{table*}
\begin{figure*}[!ht]
    \centering
    \includegraphics[width=\linewidth]{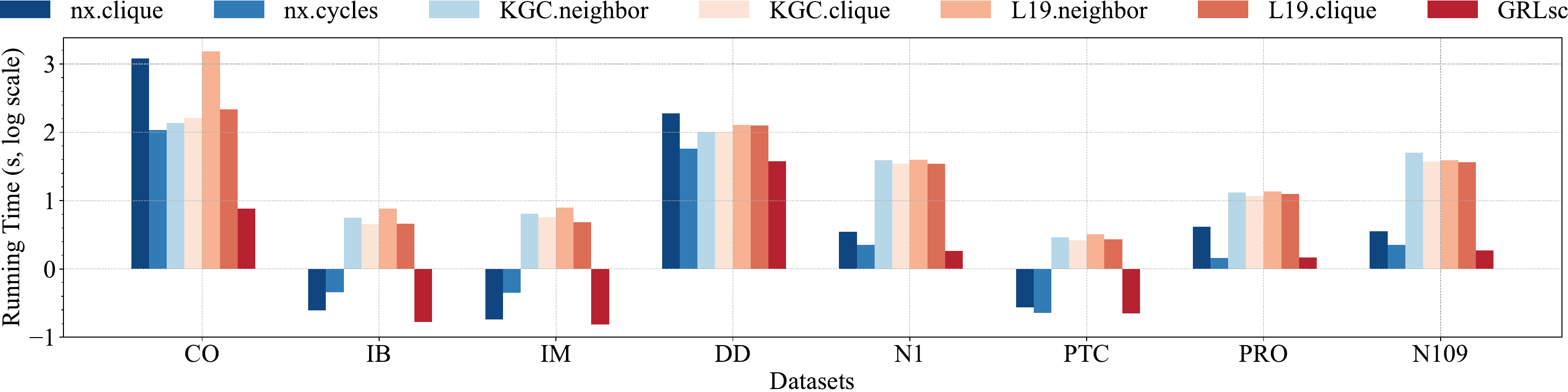}
    \caption{Runtimes of all coarsening algorithms. We make logarithmic processing due to different scales of datasets.}
    \label{fig:time}
\end{figure*}
\subsection{Coarsening Strategy Ablation}
To better discuss the impact of different coarsening strategies on the model performance, we select four categories of graph coarsening schemes: random, networkx (N)~\cite{ref:networkx}, KGC (K)~\cite{ref:KGC}, and L19 (L)~\cite{ref:L19}.
Under each category, there are two variants: neighbor (*n) and clique (*c).

In terms of implementation, we set $K=5$ and $r=0.5$, according to the paper~\cite{ref:KGC,ref:L19}. The algorithm strategy is \emph{greedy}, and other parameters are consistent with the origin.
To ensure alignment with LCC4GC, the maximum coarsening level is set to 1, which means only coarsening the original graph without higher-level compressions.
The random algorithm sets $R=K=5$, and each supernode randomly selects components from the sampling sets until all nodes are covered.
Table~\ref{tab:ablation_diff} shows the model performance under different coarsening strategies, where LCC achieves the optimal results under all eight datasets.
It indicates that the graph coarsening view obtained by LCC is more suitable for graph classification tasks.
\subsubsection{Time Analysis}
\begin{table*}[!t]
    \caption{Coarsening strategy ablation studies on all eight datasets. LC is the abbreviation of LCC.}
    \centering
    \begin{tabular}{lcccc}
        \hline
        \bf{Method} & \bf{CO} & \bf{DD} & \bf{PTC} & \bf{PRO} \\
        \hline
        Neighbor & $98.12^{0.65}$ & $98.64^{1.94}$ & $70.67 ^{6.53}$ & $75.20 ^{3.99}$ \\
        Random & $97.94^{0.58}$ & $99.32^{0.74}$ & $71.81 ^{4.61}$ & $72.51 ^{5.68}$ \\
        \hline        
        Nn & $98.02^{0.60}$ & $86.16 ^{4.10}$ & $75.92^{6.31}$ & $74.30 ^{4.57}$ \\
        Nc & $98.04^{0.58}$ & $86.19^{5.62}$ & $72.05 ^{6.67}$ & $72.77 ^{6.16}$ \\
        \hline
        Kn & $98.08^{0.58}$ & $97.71^{2.57}$ & $75.00 ^{2.66}$ & $73.13 ^{3.70}$ \\
        Kc & $96.06^{0.77}$ & $82.00 ^{7.04}$ & $70.61^{8.86}$ & $74.75 ^{5.05}$ \\
        \hline
        Ln & $96.92^{0.84}$ & $96.44 ^{4.82}$ & $74.06^{8.07}$ & $74.21 ^{4.45}$ \\
        Lc & $97.28^{0.66}$ & $91.08 ^{3.00}$ & $71.20^{6.10}$ & $74.30 ^{3.65}$ \\
        \hline
        LC & $\underline{\bf{98.18}}^{0.42}$ & $\underline{\bf{99.58}}^{0.78}$ & $\underline{\bf{76.43}}^{5.22}$ & $\underline{\textbf{76.37}}^{4.37}$ \\
        \hline
    \end{tabular}
    \begin{tabular}{lcccc}
        \hline
        \bf{Method} & \bf{IB} & \bf{IM} & \bf{N1} & \bf{N109}\\
        \hline
        Neighbor & $89.10 ^{4.09}$ & $62.80 ^{3.45}$ & $77.79 ^{1.48}$ & $76.40 ^{1.99}$ \\
        Random & $89.10^{3.78}$ & $61.70^{1.90}$ & $77.79 ^{2.36}$ & $74.10 ^{1.80}$ \\
        \hline        
        Nn & $87.00^{2.31}$ & $62.47 ^{4.06}$ & $74.84 ^{2.59}$ & $73.81^{2.21}$ \\
        Nc & $87.90 ^{2.62}$ & $62.53 ^{3.41}$ & $75.69 ^{1.92}$ & $73.88 ^{1.81}$ \\
        \hline
        Kn & $87.00 ^{1.95}$ & $58.87 ^{4.03}$ & $76.18 ^{2.12}$ & $73.42 ^{1.36}$ \\
        Kc & $89.30 ^{2.72}$ & $59.40 ^{3.66}$ & $74.60 ^{1.91}$ & $72.84^{1.72}$ \\
        \hline
        Ln & $87.10 ^{2.62}$ & $61.70 ^{3.17}$ & $75.94 ^{2.20}$ & $73.81^{2.14}$ \\
        Lc & $88.20 ^{2.79}$ & $61.87 ^{3.55}$ & $77.81 ^{1.48}$ & $75.14^{1.86}$ \\
        \hline
        LC & $\underline{\bf{89.40}}^{2.50}$ & $\underline{\bf{63.20}}^{3.55}$ & $\underline{\textbf{79.08}}^{1.92}$ & $\underline{\textbf{76.98}}^{1.90}$ \\
        \hline
    \end{tabular}
    \label{tab:ablation_diff}
\end{table*}
We analyze the linear time complexity of LCC in section 4.2.4.
Here, we compare it with other methods.
Figure~\ref{fig:time} shows the results.
The results show that the runtime of LCC is significantly lower than other methods.
The runtime of LCC is significantly lower than other methods.
Except for CO and DD, the computation time is within 2s for all other datasets.
Runtime for CO is 7.60s, while for DD is 37.58s.
The average runtime of \emph{KGC.neighbor} is 44.55s, and \emph{L19.clique} is 54.98s.
Both of them are higher than LCC.
\begin{table*}[!hpt]
    \centering
    \caption{Scale analysis of graph coarsening. Each group represents the size of the intermediate coarsening graph under different strategies, where $\Bar{V}$ and $\Bar{E}$ represent the average nodes and edges, and $r_V$ and $r_E$ represent the scaling ratio of the coarsening graph compared to the original graph. The negative number means reducing the scale from the original graph, and the positive means expanding. O is the original graphs.}
    \begin{tabular}{lccrrrrrrrr}
        \hline
        \multicolumn{3}{l}{} & \bf{CO} & \bf{IB} & \bf{IM} & \bf{DD} & \bf{N1} & \bf{PTC} & \bf{PRO} & \bf{N109} \\
        \hline    
        \multicolumn{2}{l}{\multirow{2}{*}{O}} & $\Bar{V}$ & 74.49 & 19.77 & 13.00 & 2.8e2 & 39.06 & 25.56 & 29.87 & 29.68 \\
        && $\Bar{E}$ & 2.5e3 & 96.53 & 65.94 & 7.2e2 & 72.82 & 25.96 & 32.30 & 32.13 \\
        \hline
        \multicolumn{2}{l}{\multirow{4}{*}{Nn}} & $\Bar{V}$ & 74.49 & 19.77 & 13.00 & 2.8e2 & 29.84 & 25.30 & 39.06 & 29.65 \\
        && $r_V$ & 0.00 & 0.00 & 0.00 & 0.00 & -0.24 & -0.01 & +0.31 & 0.00 \\
        && $\Bar{E}$ & 2.5e3 & 1.0e2 & 67.94 & 7.2e2 & 42.09 & 47.85 & 77.02 & 41.98 \\
        && $r_E$ & +0.01 & +0.05 & +0.03 & +0.01 & -0.42 & +0.84 & +1.38 & +0.31 \\
        \hline
        \multicolumn{2}{l}{\multirow{4}{*}{Nc}} & $\Bar{V}$ & 42.46 & 3.77 & 2.05 & 3.0e2 & 32.31 & 25.86 & 40.79 & 32.16 \\
        && $r_V$ & -0.43 & -0.81 & -0.84 & +0.05 & -0.17 & +0.01 & +0.37 & +0.08 \\
        && $\Bar{E}$ & 7.1e3 & 11.96 & 2.81 & 2.8e3 & 1.1e2 & 1.0e2 & 2.6e2 & 1.1e2 \\
        && $r_E$ & +1.88 & -0.88 & -0.96 & +2.90 & +0.50 & +2.96 & +7.03 & +2.39 \\
        \hline
        \multicolumn{2}{l}{\multirow{4}{*}{Kn}} & $\Bar{V}$ & 5.24 & 2.05 & 1.51 & 90.15 & 13.86 & 12.49 & 13.31 & 13.78 \\
        && $r_V$ & -0.93 & -0.90 & -0.88 & -0.68 & -0.65 & -0.51 & -0.55 & -0.54 \\
        && $\Bar{E}$ & 19.17 & 1.47 & 0.64 & 4.5e2 & 24.53 & 18.66 & 37.13 & 24.37 \\
        && $r_E$ & -0.99 & -0.98 & -0.99 & -0.37 & -0.66 & -0.28 & +0.15 & -0.24 \\
        \hline
        \multicolumn{2}{l}{\multirow{4}{*}{Kc}} & $\Bar{V}$ & 10.91 & 3.26 & 1.92 & 1.2e2 & 17.33 & 16.31 & 17.74 & 17.27 \\
        && $r_V$ & -0.85 & -0.84 & -0.85 & -0.59 & -0.56 & -0.36 & -0.41 & -0.42 \\
        && $\Bar{E}$ & 60.55 & 5.38 & 1.80 & 5.1e2 & 27.54 & 22.76 & 46.28 & 27.41 \\
        && $r_E$ & -0.98 & -0.94 & -0.97 & -0.29 & -0.62 & -0.12 & +0.43 & -0.15 \\
        \hline
        \multicolumn{2}{l}{\multirow{4}{*}{Ln}} & $\Bar{V}$ & 46.30 & 12.46 & 9.83 & 1.4e2 & 15.27 & 13.67 & 19.88 & 15.17 \\
        && $r_V$ & -0.38 & -0.37 & -0.24 & -0.50 & -0.61 & -0.47 & -0.33 & -0.49 \\
        && $\Bar{E}$ & 1.3e3 & 57.22 & 46.81 & 5.2e2 & 25.57 & 21.26 & 47.52 & 25.40 \\
        && $r_E$ & -0.48 & -0.41 & -0.29 & -0.28 & -0.65 & -0.18 & +0.47 & -0.21 \\
        \hline
        \multicolumn{2}{l}{\multirow{4}{*}{Lc}} & $\Bar{V}$ & 46.12 & 12.42 & 9.76 & 1.4e2 & 17.35 & 16.00 & 20.25 & 17.28 \\
        && $r_V$ & -0.38 & -0.37 & -0.25 & -0.50 & -0.56 & -0.37 & -0.32 & -0.42 \\
        && $\Bar{E}$ & 1.3e3 & 57.92 & 45.58 & 5.4e2 & 27.55 & 22.83 & 50.80 & 27.43 \\
        && $r_E$ & -0.47 & -0.40 & -0.31 & -0.25 & -0.62 & -0.12 & +0.57 & -0.15 \\
        \hline
        \multicolumn{2}{l}{\multirow{4}{*}{LC}} & $\Bar{V}$ & 12.52 & 3.45 & 2.01 & 1.6e2 & 18.88 & 20.49 & 25.85 & 18.73 \\
        && $r_V$ & -0.83 & -0.83 & -0.85 & -0.45 & -0.52 & -0.20 & -0.13 & -0.37 \\
        && $\Bar{E}$ & 29.38 & 3.07 & 1.22 & 3.1e2 & 18.97 & 20.07 & 40.54 & 18.85 \\
        && $r_E$ & -0.99 & -0.97 & -0.98 & -0.57 & -0.74 & -0.23 & +0.26 & -0.41 \\
        \hline
    \end{tabular}
    \label{tab:scale}
\end{table*}
\subsubsection{Scale Analysis}
Though we do not require LCC to achieve the optimal coarsening effect, we still analyze and compare the scale of the intermediate coarsening graph $G_{lcc}$.\
We experiment with three other categories of strategies and their variants in addition to random algorithms.
Given a set of original input graphs $\mathcal{G}_o=\{G_o^1, \cdots, G_o^M\}$, we calculate and collate the scale of $\mathcal{G}_{lcc}=\{G_{lcc}^1, \cdots, G_{lcc}^M\}$.
We show average node and edge numbers through $\Bar{V} = \frac{1}{M}\sum_i\mathcal{V}_{G_i}$ and $\Bar{E} = \frac{1}{M}\sum_i\mathcal{E}_{G_i}$, and analyze the scale changes leveraging formula $r_V = \frac{\Bar{V}_{lcc}-\Bar{V}_o}{\Bar{V}_o}$ and $r_E = \frac{\Bar{E}_{lcc}-\Bar{E}_o}{\Bar{E}_o}$.
Table~\ref{tab:scale} shows the results.

In general, the LCC we designed can achieve a considerable coarsening result.
The coarsening rate at the node level is about 52.2\%.
It is slightly higher at the edge level, 57.9\% or so.
In baselines, the best one is \emph{KGC.neighbor}, with a 70.5\% compression rate at the node level and 54.7\% at the edge level.
We do not expect LCC to achieve similar high rates.
When the coarsening process is to the maximum extent, amounts of high-level structural information are lost.
As the number of nodes decreases linearly, edges have a faster decay.
We cannot mine enough structural information in very simple hypergraphs.
Thus, LCC achieves the maximum balance between coarsening nodes and preserving structures.
\subsubsection{Visualization}
\begin{figure*}[!ht]
    \centering
    \subfloat[Cliques: IB (14)]{
        \includegraphics[width=.95\linewidth]{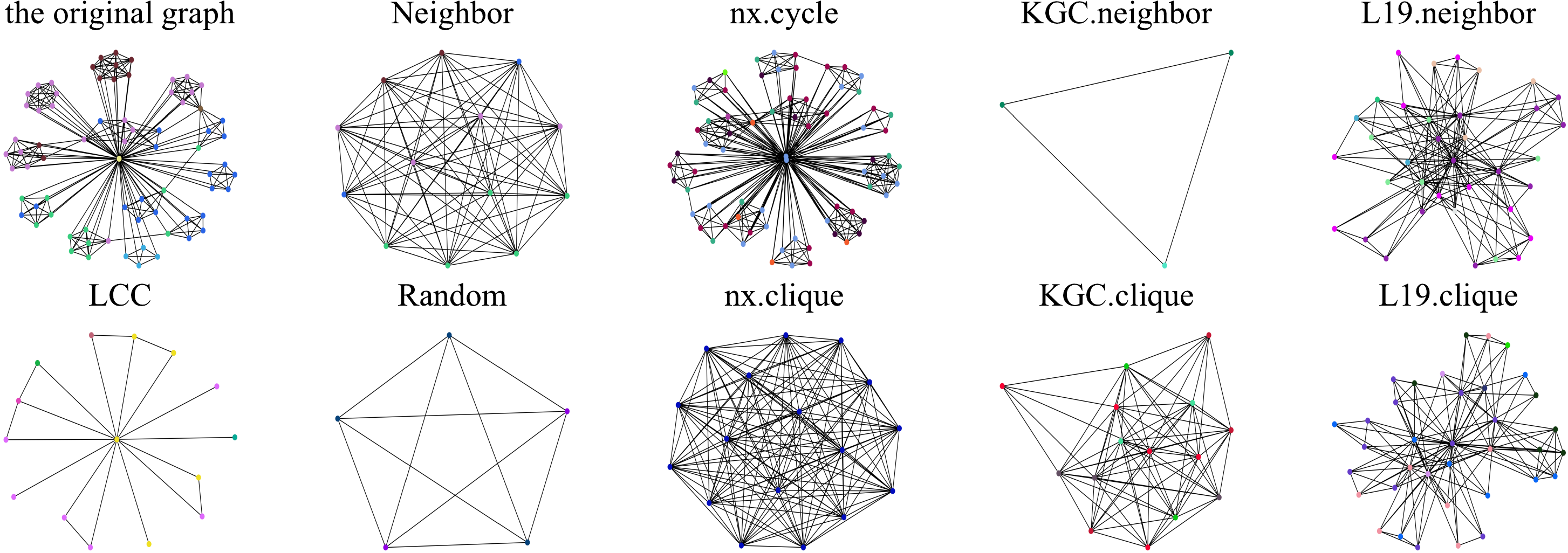}
        \label{fig:CSN109}
    }
    \hfill
    \subfloat[Loops: N109 (77)]{
        \includegraphics[width=.95\linewidth]{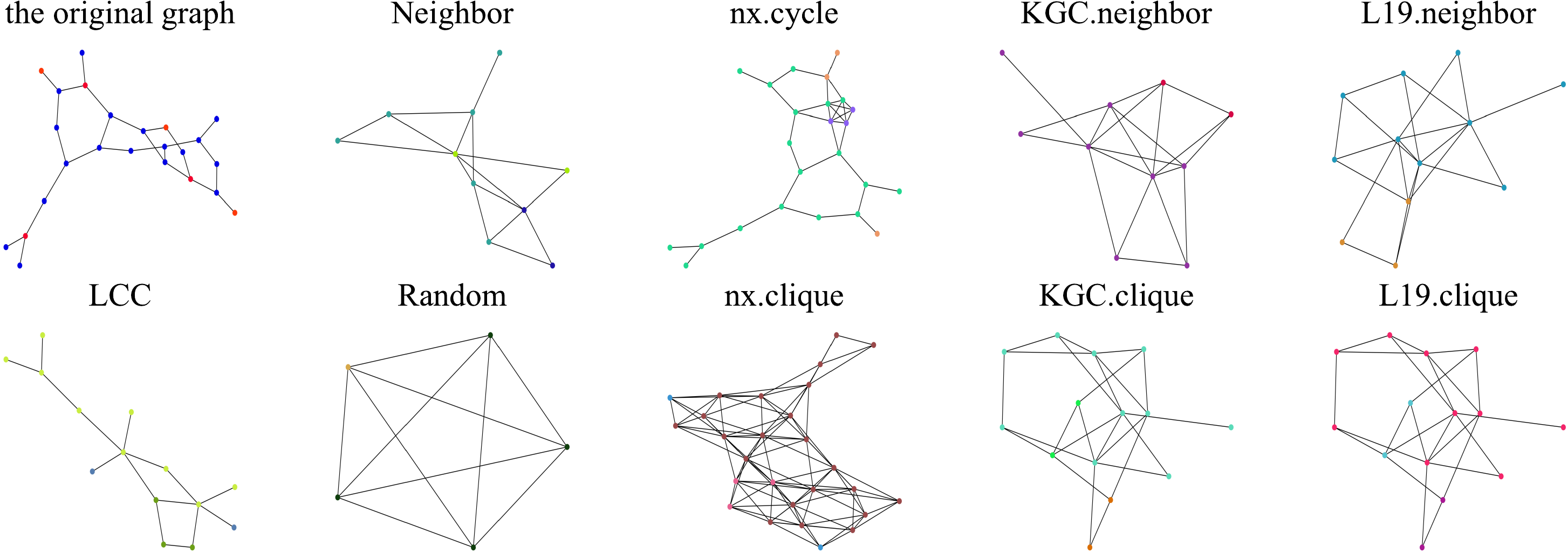}
        \label{fig:CSN109}
    }
    \hfill
    \caption{Visualization of different coarsening strategies. In each group, row 1 col 1 is the original input graph, row 2 col 1 is the coarsening result of LCC4GC, and the other columns are two variants of the method in the same category. Different node colors represent different node classes. The graph layout does not contain semantics such as rotation and symmetry but only shows the connections.}
    \label{fig:co_case}
\end{figure*}
We further analyze the coarsening comparison using visualization.
Figure~\ref{fig:co_case} shows the results.
It appears that LCC4GC can obtain the coarsening graph representation closest to the original graph.
We retain the connections between supernodes to the greatest extent and mine the high-level structural information thoroughly based on the coarsening procedures.
Other algorithms lack balance in coarsening and structural representation.
Random algorithms break away from the original structural semantics and pursue to cover the whole graph by specifying the number of nodes.
KGC presents a more concise coarsening pattern.
However, it fails to express high-level structural information and has poor adaptability to loops.
L19 relaxes the restriction, not limiting nodes located in multiple partitioning sets, but still cannot refine the structures.

\subsection{Case Study}
\begin{figure*}[!hptb]
    \centering
    \subfloat[CO (5)]{
        \includegraphics[width=.45\linewidth]{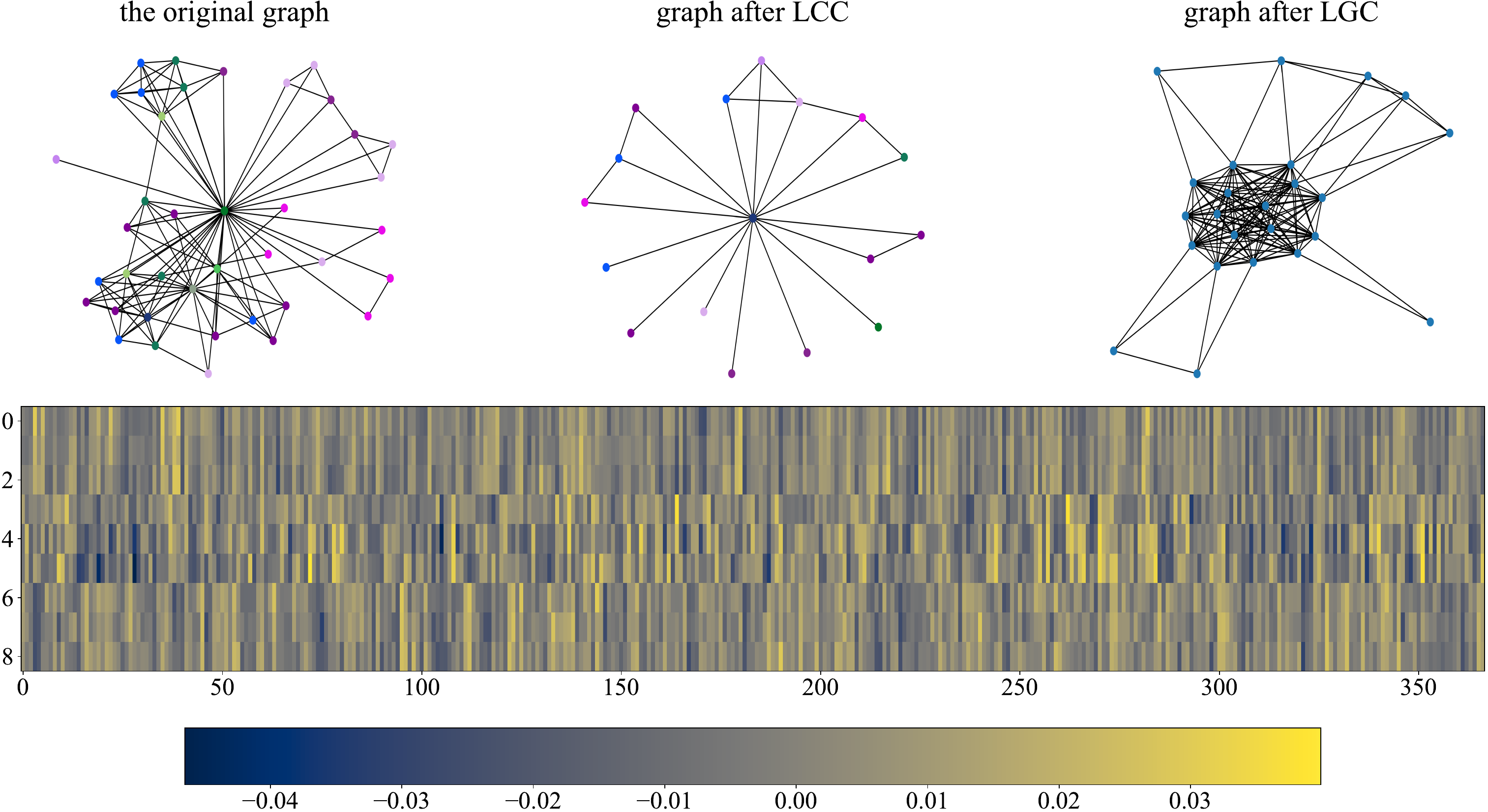}
    }
    \subfloat[DD (0)]{
        \includegraphics[width=.45\linewidth]{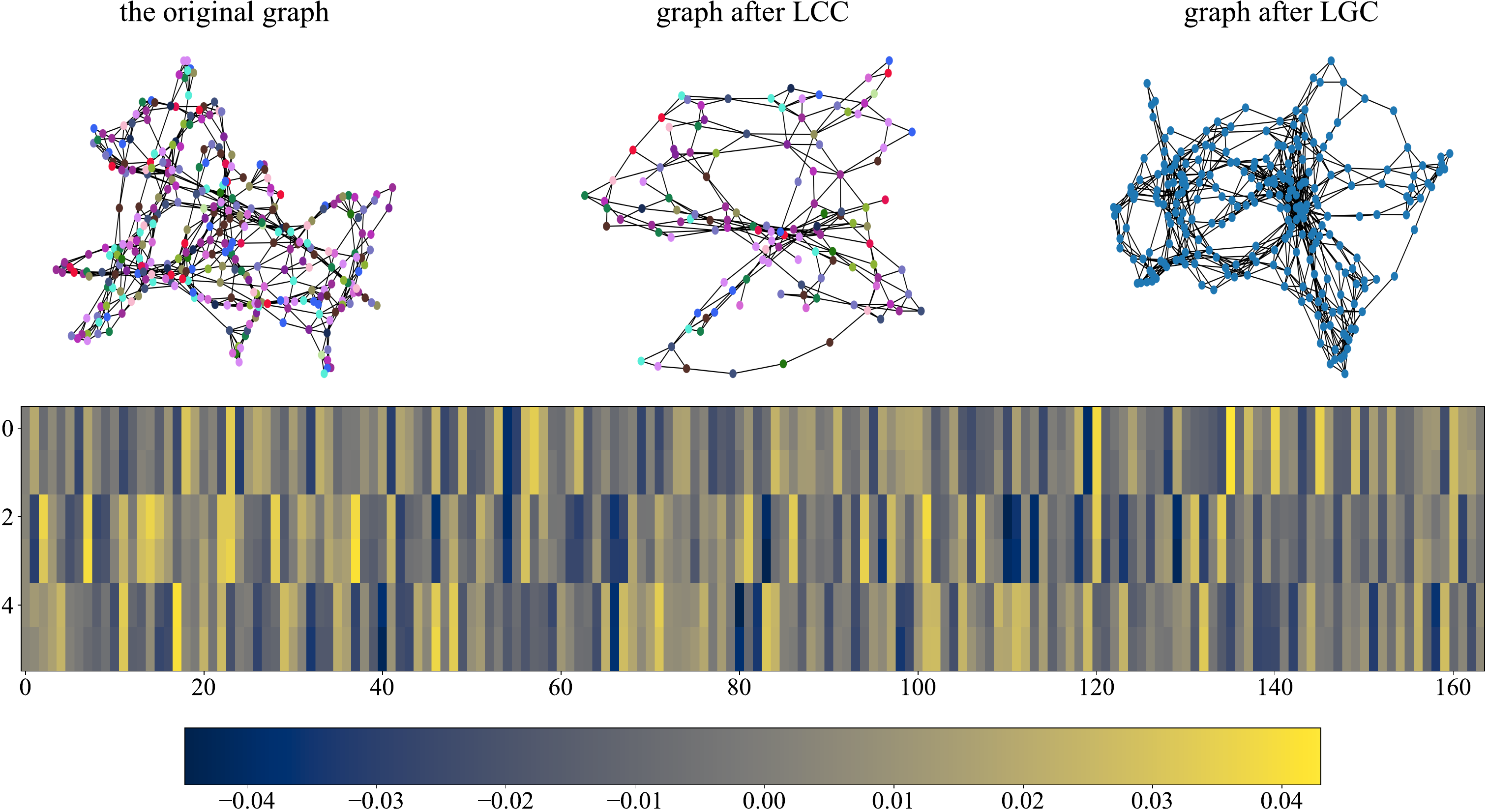}
    }
    \hfill
    \subfloat[IB (5)]{
        \includegraphics[width=.45\linewidth]{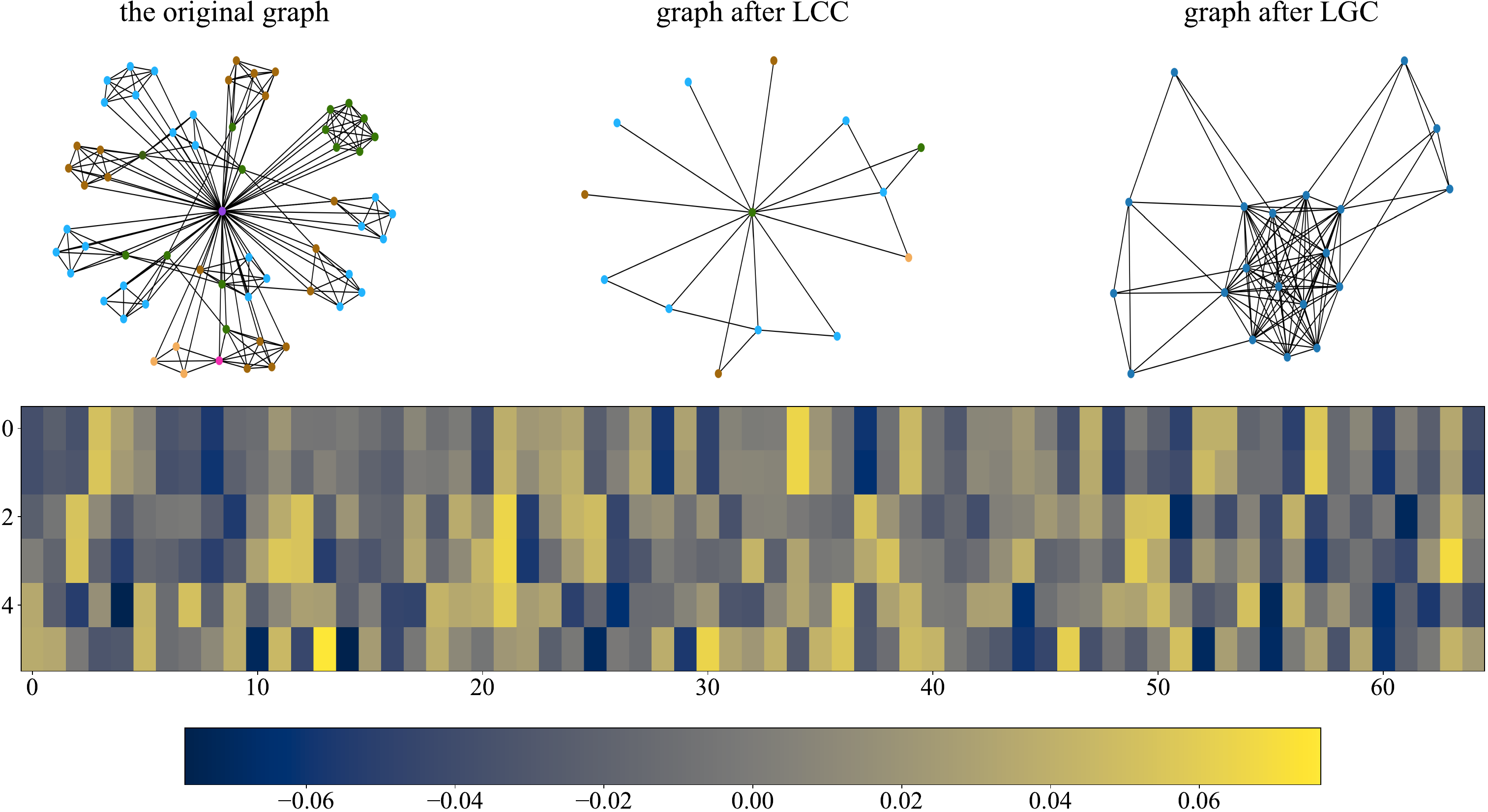}
    }
    \subfloat[IM (268)]{
        \includegraphics[width=.45\linewidth]{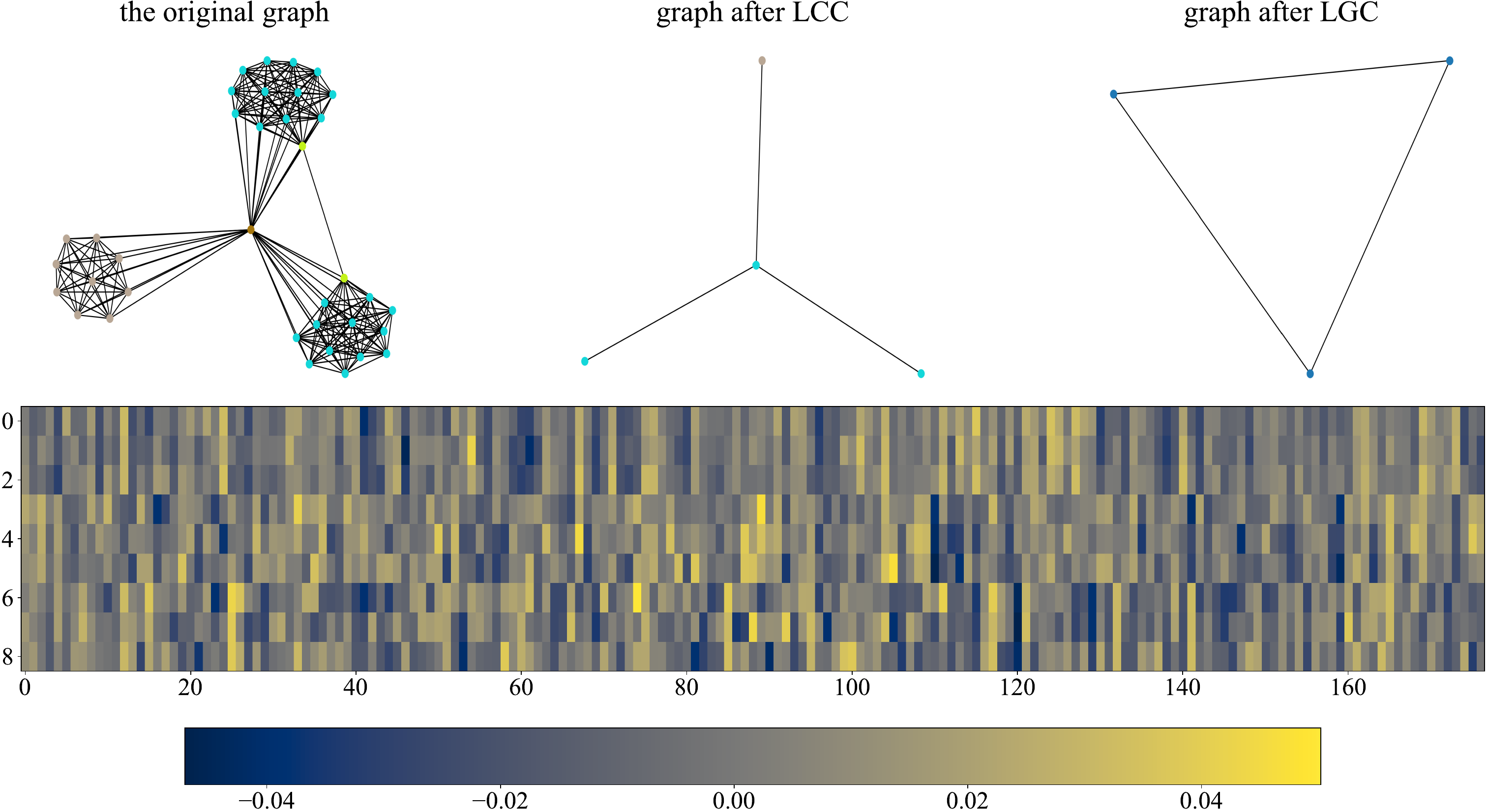}
    }
    \hfill
    \subfloat[N1 (45)]{
        \includegraphics[width=.45\linewidth]{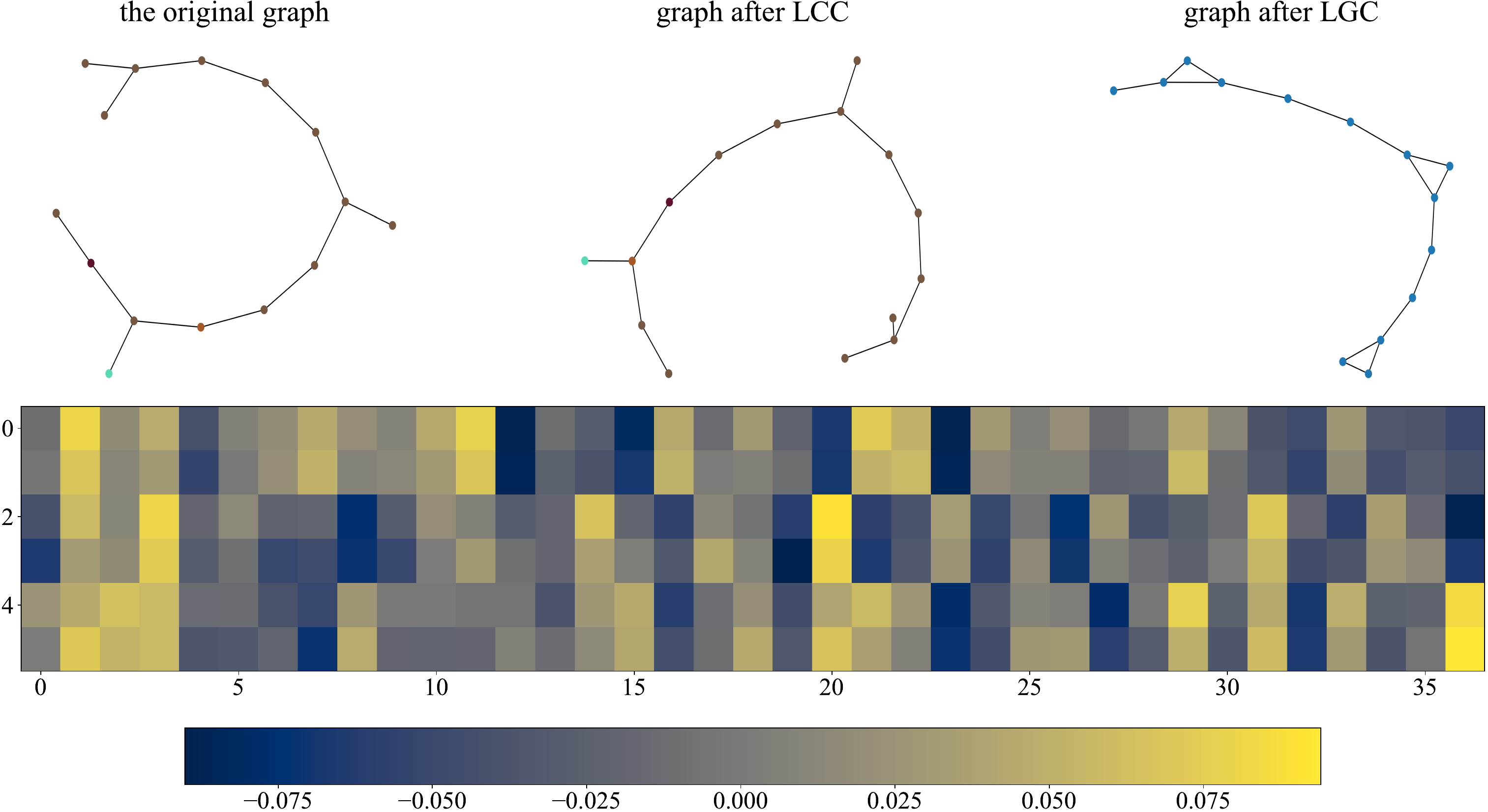}
    }
    \subfloat[N109 (540)]{
        \includegraphics[width=.45\linewidth]{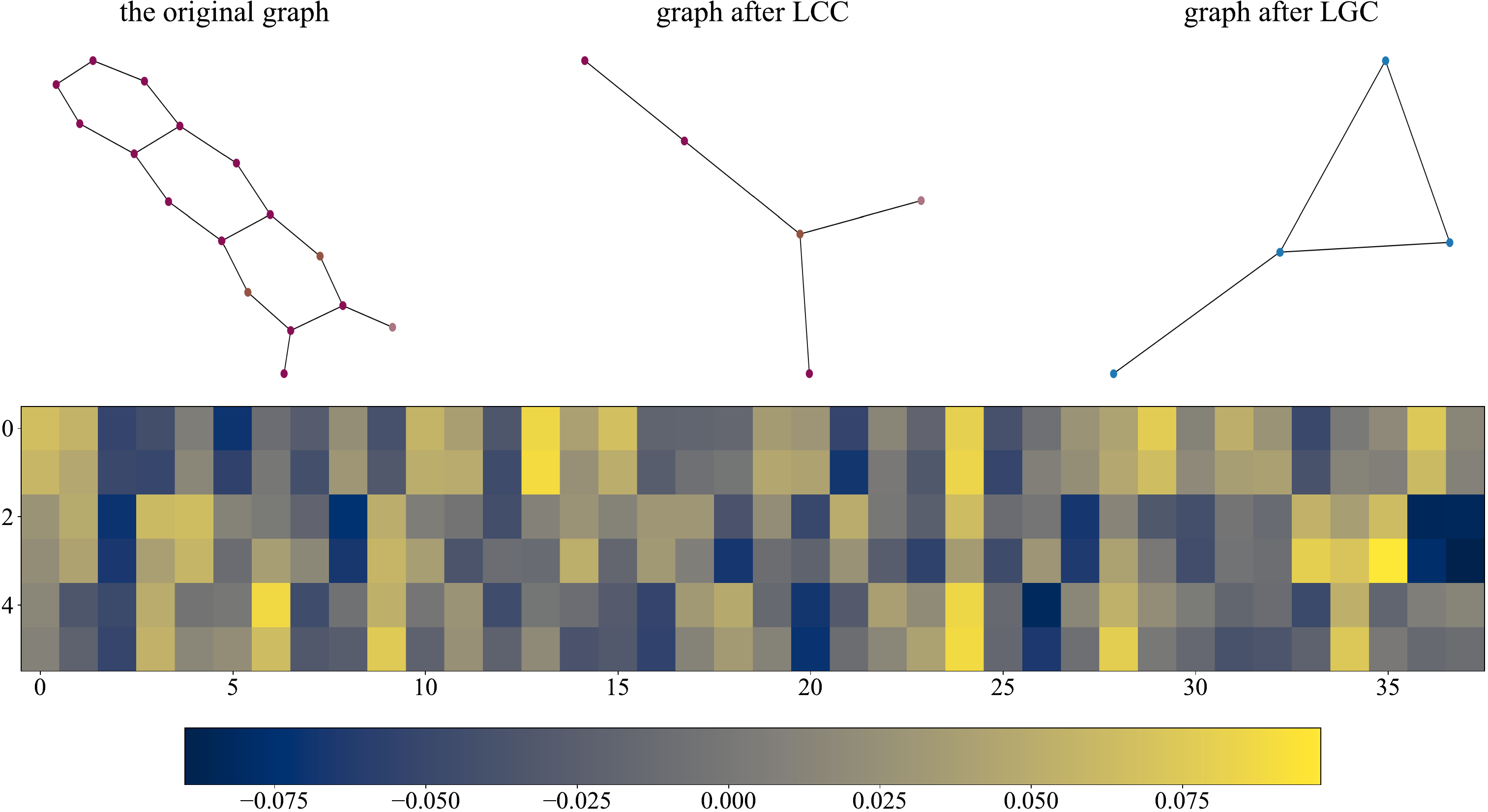}
    }
    \hfill
    \subfloat[PTC (7)]{
        \includegraphics[width=.45\linewidth]{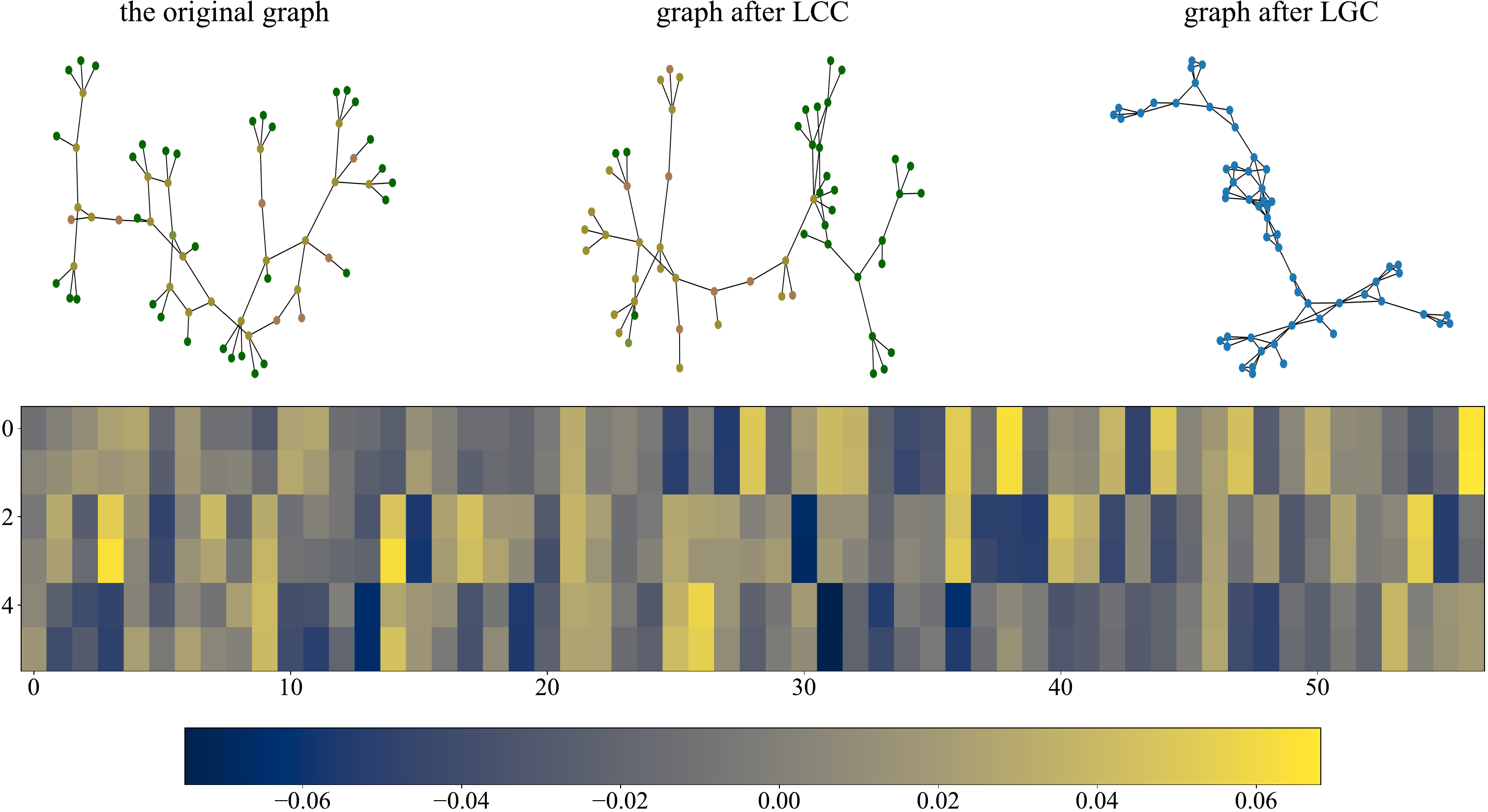}
    }
    \subfloat[PRO (6)]{
        \includegraphics[width=.45\linewidth]{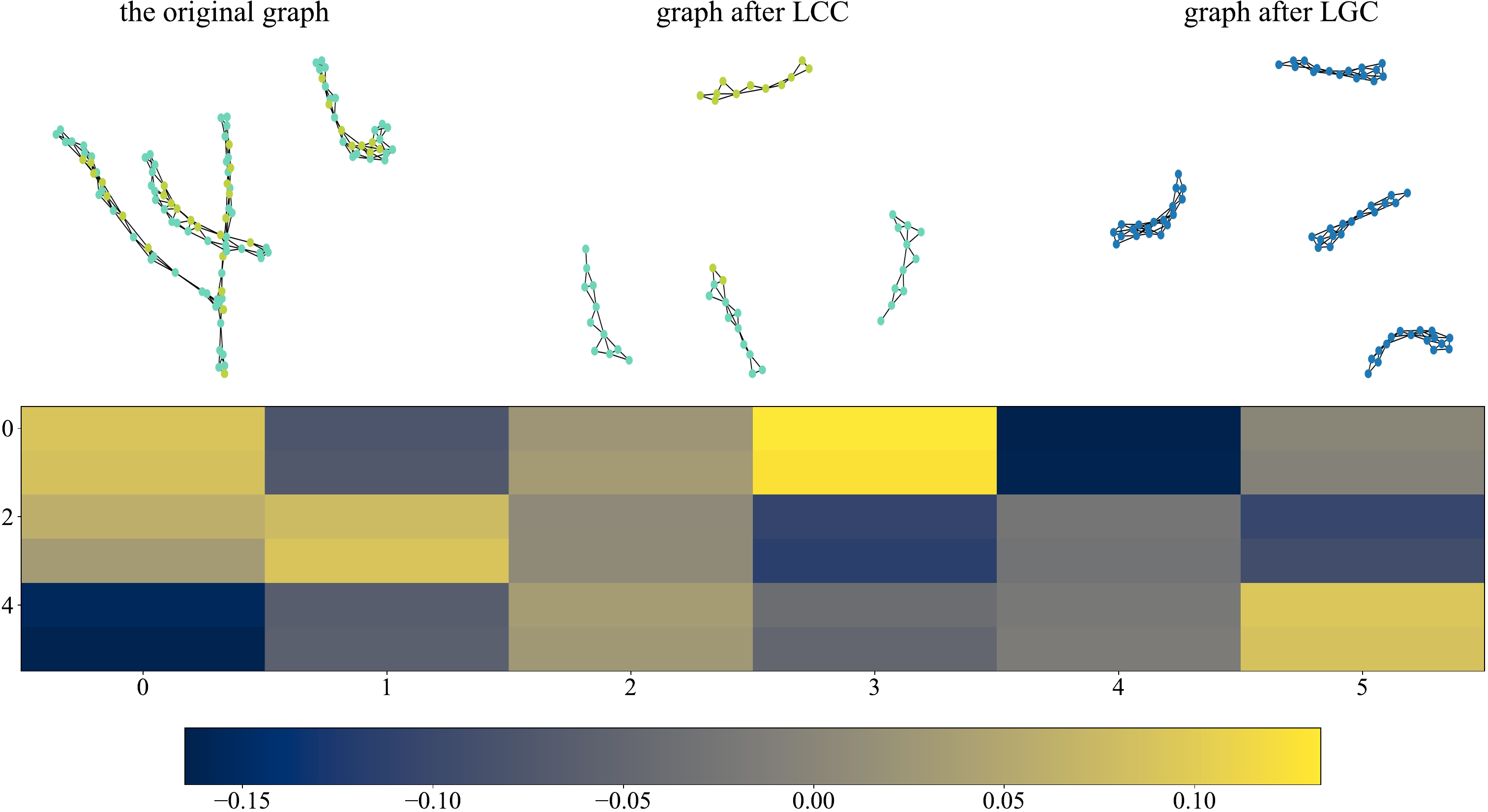}
    }
    \caption{Case study for all datasets. Each case contains three graphs from left to right: the original graph view, the graph coarsening view, and the line graph conversion view. The heat plot below them is the classifier weights, with the x-axis representing the feature dimension and the y-axis representing the three graphs (each graph view takes rows of the number of graph classes).}
    \label{fig:case}
\end{figure*}
We conduct case studies on all eight datasets to fully cover each category and explore the choices of the classifier after LCC4GC.
Figure~\ref{fig:case} illustrates the results.

LCC4GC focuses on loops and cliques and thus performs well on datasets rich in such structures.
As for loops, we take (f) N109 as an example, which contains three benzene rings connected in turn.
LCC can identify such loop structures and coarsen them into three supernodes.
The remaining two independent carbon atoms are then converted by LGC into a triangle, as shown in the figure.
We can see from the heatmap that both LCC and LGC contain vital dimensions contributing most to the classification weights, where LCC in (35, 3), LGC in (6, 4), (24, 5), etc.

As for cliques, we take (c) IB as an example.
It forms 12 independent cliques centered on the actor or actress represented by the purple node.
A clique represents a collection of actors or actresses from one scene, and some actors or actresses may appear in more than one scene.
LCC weakens the concept of nodes but highlights the clique structure by coarsening, strengthening the connection between scenes with corporate actors or actresses.
LGC further emphasizes the structural information of the original graph.
A clique under the edge-centric view shows a new representation of the purple center node.
Compared to the previous case, for datasets with prominent clique structures, LCC contributes more to the weight than LGC, where LCC in (2, 2), (21, 3), (63, 3), and LGC in (13, 5), (46, 5), and so on.


Finally, we can still find in the heatmap that the classifier retains weights for some feature dimensions in the original graph view.
It is because, after the constrained coarsening and conversion, some nodes retained in the graph still hold valuable structural information, such as the bridge node connecting two structures and the edge node connecting inside and outside a clique.
They should be paid equal attention to those unique components.

\section{Conclusion}
In this paper, we propose an efficient \textbf{L}oop and \textbf{C}lique \textbf{C}oarsening algorithm with linear complexity \textbf{for} \textbf{G}raph \textbf{C}lassification (LCC4GC) on GT architectures.
We focus on loops and cliques and investigate the feasibility of coarsening particular structures into hypernodes.
We build three unique views via graph coarsening and line graph conversion, which helps to learn high-level structural information and strengthen relative position information.
We evaluate the performance of LCC4GC on eight real-world datasets, and the experimental results demonstrate that LCC4GC can outperform SOTAs in graph classification task.
Though LCC4GC achieves remarkable results, we still have a long way to go.
In the future, there are two main directions for discussion.
First, graphs in the real world are constantly changing over time, we will try to introduce dynamic graphs.
Second, graph structures are more complex and diverse than just loops and cliques, we will consider extending to general structures to mine richer information at a high level.

\bibliographystyle{elsarticle-num} 
\bibliography{main}





\end{document}